\documentclass[10pt,twocolumn,letterpaper]{article}

\usepackage{wacv}              

\usepackage{graphicx}
\usepackage{amsmath}
\usepackage{amssymb}
\usepackage{booktabs}

\usepackage[pagebackref,breaklinks,colorlinks]{hyperref}

\usepackage{booktabs} 
\usepackage{subcaption}
\usepackage{bbm}
\usepackage{arydshln}
\usepackage{xcolor}
\usepackage{multirow}
\usepackage[algo2e,ruled,vlined,linesnumbered]{algorithm2e}
\usepackage{amsfonts}
\usepackage{tikz}
\usepackage{comment}
\usepackage{dsfont}
\usepackage{color, colortbl}
\usepackage{xcolor}
\usepackage{enumitem}

\usepackage{amsmath,amsfonts,bm}

\def\eqref#1{equation~\ref{#1}}

\def\1{\bm{1}}

\DeclareMathAlphabet{\mathsfit}{\encodingdefault}{\sfdefault}{m}{sl}
\SetMathAlphabet{\mathsfit}{bold}{\encodingdefault}{\sfdefault}{bx}{n}

\DeclareMathOperator*{\argmax}{arg\,max}

\definecolor{Gray}{gray}{0.9}
\usepackage{graphicx}
\usepackage{multirow}
\usepackage{multicol}
\usepackage{xspace}
\usepackage{booktabs}
\usepackage{tabularx,  makecell, caption}
\usepackage{diagbox}
\usepackage{algorithmic}
\usepackage{algorithm}

\usepackage[capitalize]{cleveref}
\crefname{section}{Sec.}{Secs.}
\Crefname{section}{Section}{Sections}
\Crefname{table}{Table}{Tables}
\crefname{table}{Tab.}{Tabs.}

\def\wacvPaperID{2124} 
\def\confName{WACV}
\def\confYear{2025}

\begin{document}

\title{Re-evaluating Group Robustness via Adaptive Class-Specific Scaling}

\author{
Seonguk Seo$^1$ \qquad Bohyung Han$^{1,2}$ \\
$^1$ECE \& $^{2}$IPAI, Seoul National University\\
 {\tt\small \{seonguk, bhhan\}@snu.ac.kr}
}
\maketitle

\begin{abstract}
Group distributionally robust optimization, which aims to improve robust accuracies---worst-group and unbiased accuracies---is a prominent algorithm used to mitigate spurious correlations and address dataset bias. 
Although existing approaches have reported improvements in robust accuracies, these gains often come at the cost of average accuracy due to inherent trade-offs. 
To control this trade-off flexibly and efficiently, we propose a simple class-specific scaling strategy, directly applicable to existing debiasing algorithms with no additional training.
We further develop an instance-wise adaptive scaling technique to alleviate this trade-off, even leading to improvements in both robust and average accuracies. 
Our approach reveals that a na\"ive ERM baseline matches or even outperforms the recent debiasing methods by simply adopting the class-specific scaling technique.
Additionally, we introduce a novel unified metric that quantifies the trade-off between the two accuracies as a scalar value, allowing for a comprehensive evaluation of existing algorithms. 
By tackling the inherent trade-off and offering a performance landscape, our approach provides valuable insights into robust techniques beyond just robust accuracy. 
We validate the effectiveness of our framework through experiments across datasets in computer vision and natural language processing domains.
\end{abstract}


\section{Introduction}
\label{sec:intro}

\begin{figure}[t]
\centering
    \begin{subfigure}[m]{1\linewidth}
    	\includegraphics[width=\linewidth]{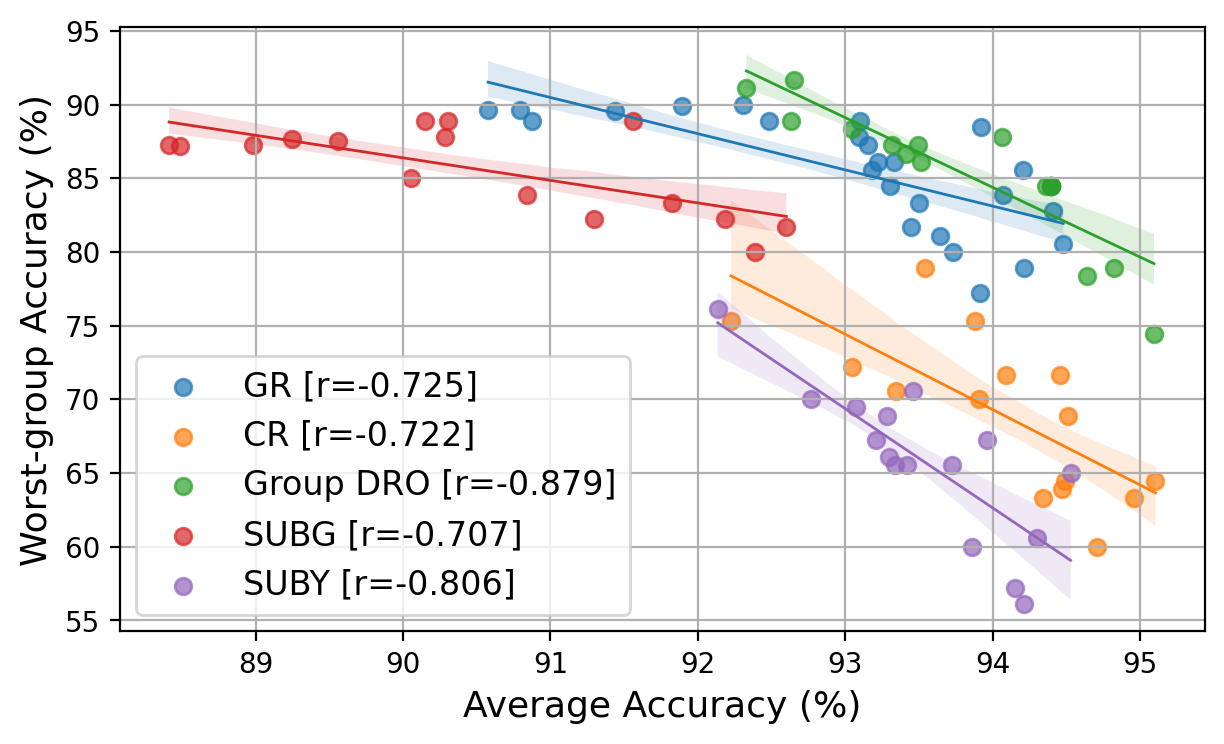}\vspace{-0mm}
	\label{fig:observation_worst}
    \end{subfigure} 
    \vspace{-2mm}
    \caption{The scatter plots illustrate trade-offs between robust and average accuracies of existing algorithms with ResNet-18 on CelebA.
    We visualize the results from multiple runs of each algorithm and present the relationship between the two accuracies.
    The lines denote the linear regression results of individual algorithms and $r$ in the legend indicates the Pearson coefficient correlation.
    }
    \label{fig:observation_tradeoff}
\end{figure}
\begin{figure*}[t]
\centering
    \begin{subfigure}[m]{0.45\linewidth}
    	\includegraphics[width=\linewidth]{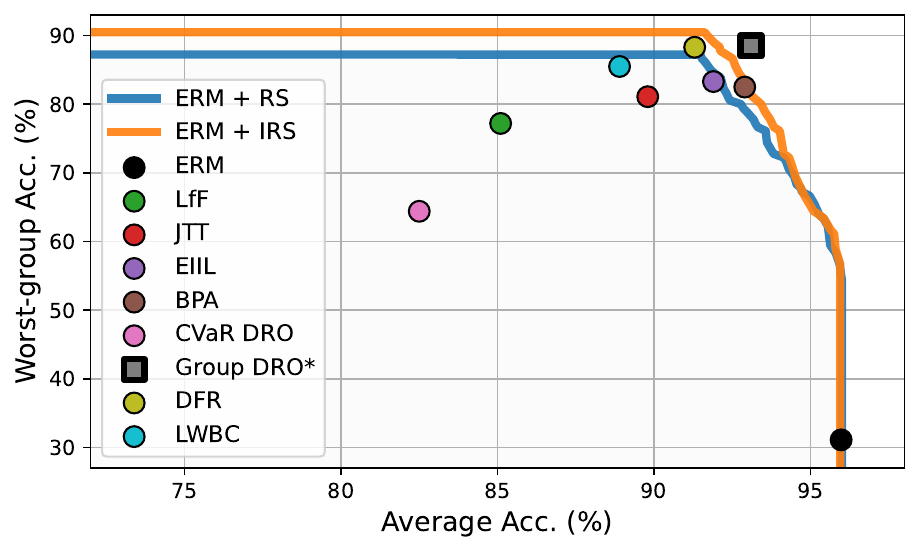} \vspace{-5mm}
	\subcaption{Worst-group accuracy}
	\label{fig:teaser_worst}
    \end{subfigure}     
    	\hspace{10mm}
        \begin{subfigure}[m]{0.45\linewidth}
    	\includegraphics[width=\linewidth]{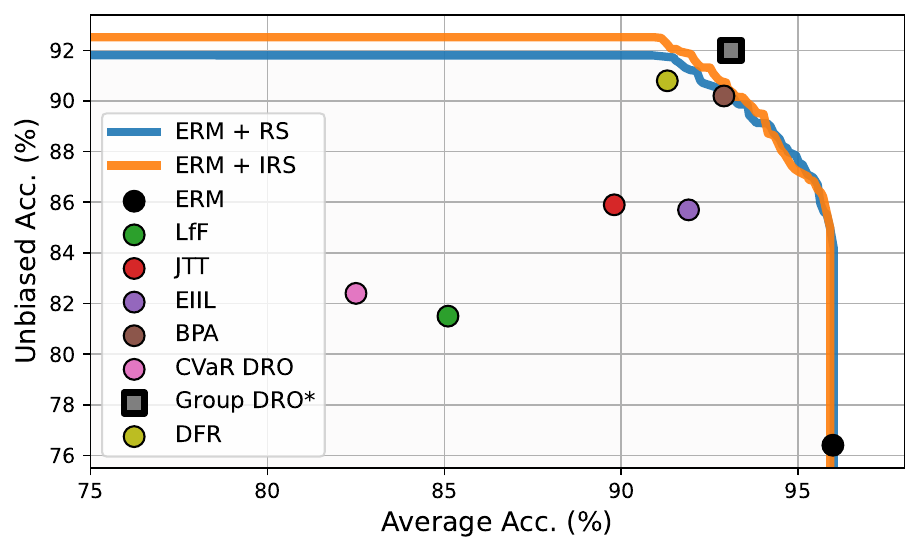} \vspace{-5mm}
	\subcaption{Unbiased accuracy}
	\label{fig:teaser_unbias}
	\end{subfigure} 
    \caption{Comparison between the baseline ERM and existing debiasing approaches with ResNet-50 on CelebA.
    Existing works have improved robust accuracy substantially compared to ERM, but our robust scaling strategies such as RS and IRS enable ERM to catch up with or even outperform them without further training.
    }
    \label{fig:teaser}
\end{figure*}

Machine learning models achieve remarkable performance across various tasks via empirical risk minimization (ERM).
However, they are often vulnerable to spurious correlations and dataset biases, resulting in poor classification performance for minority groups despite high average accuracy.
For example, in the Colored MNIST dataset~\cite{IRM, ReBias}, a strong correlation exists between digit labels and foreground colors. 
Consequently, trained models tend to rely on these unintended patterns, resulting in significant performance degradation when classifying digits with rare color associations that are underrepresented in the training data.

Since spurious correlations are well-known to degrade generalization performance in minority groups, group distributionally robust optimization~\cite{GroupDRO} has been widely adopted to address algorithmic bias. 
Numerous approaches~\cite{huang2016learning, GroupDRO, seo2021unsupervised, LfF, sohoni2020no, levy2020large, JTT} have achieved high robust accuracies, such as worst-group or unbiased accuracies, across various tasks and datasets. 
However, despite these improvements, they often come at the expense of average accuracy, and little effort has been made to comprehensively evaluate the robust and average accuracies together.
Figure~\ref{fig:observation_tradeoff} demonstrates the trade-offs of existing algorithms.

This paper addresses the limitations of current research trends by introducing a simple post-processing technique, \textit{robust scaling}, which efficiently performs class-specific scaling on prediction scores and conveniently controls the trade-off between robust and average accuracies at test time.
It allows us to identify any desired performance points under various metrics such as average accuracy, unbiased accuracy, worst-group accuracy, or balanced accuracy, along the accuracy trade-off curve derived from a pretrained model with negligible extra computational overhead.
The proposed robust-scaling method can be seamlessly plugged into various existing debiasing algorithms to improve target objectives within the trade-off.

An interesting observation is that, by adopting the proposed robust scaling, even the ERM baseline accomplishes competitive performance without extra training compared to the recent group distributionally robust optimization approaches~\cite{JTT, LfF, GroupDRO, kim2022learning, seo2021unsupervised, creager2021environment, levy2020large, kirichenko2022last, zhang2022correct}, as illustrated in Figure~\ref{fig:teaser}.
Furthermore, we propose an advanced robust scaling algorithm that adaptively applies scaling to individual examples adaptively based on their cluster membership at test time to maximize performance.
This instance-wise adaptive scaling strategy effectively mitigates the trade-off and delivers performance improvements in both robust and average accuracies.

By taking advantage of the robust scaling technique, we develop a novel comprehensive evaluation metric that consolidates insights into the trade-off of group robustness algorithms, providing a unique perspective on group distributionally robust optimization. 
We argue that assessing robust accuracy in isolation, without accounting for average accuracy, provides an incomplete picture and a unified evaluation of debiasing algorithms is required.
For a comprehensive performance evaluation, we introduce \textit{robust coverage}, a new measure that effectively captures the trade-off between average and robust accuracies from a Pareto optimal perspective, summarizing each algorithm's performance with a single scalar value.

\vspace{-2mm}
\paragraph{Contribution} 
We propose a simple yet effective approach for group robustness by analyzing the trade-off between robust and average accuracies.
Our framework captures the complete landscape of robust-average accuracy trade-offs, facilitates understanding the behavior of existing debiasing techniques, and enables optimization of arbitrary objectives along the trade-off curve without additional training.
We emphasize that our framework does not solely focus on performance improvement in robust accuracy; more importantly, \textbf{our method not only highlights the inherent trade-offs in existing debiasing approaches but also facilitates the identification of desired performance points based on target objectives, paving the way for accurate, fair, and comprehensive evaluations of group robustness.}
%
Our main contributions are summarized as follows.%

 \begin{itemize} 
     \item[$\bullet$] We propose a training-free class-specific scaling strategy to capture and control the trade-off between robust and average accuracy with negligible computational cost. 
     This approach allows us to optimize a debiasing algorithm towards arbitrary objectives within the trade-off, building on top of any existing models. 

    \item[$\bullet$] We develop an instance-wise robust scaling algorithm by extending the original class-specific scaling with joint consideration of feature clusters. This technique is effective to alleviate the trade-off and improve both robust and average accuracy. 
    
    \item[$\bullet$] We introduce a novel comprehensive and unified performance evaluation metric based on the robust scaling method, which summarizes the trade-off as a scalar value from the Pareto optimal perspective. 

    \item[$\bullet$] The extensive experiments analyze the characteristics of existing methods and validate the effectiveness of our frameworks on the multiple standard benchmarks.
\end{itemize}



\section{Related Works}
\label{sec:related}

Mitigating spurious correlation has emerged as an important problem in many areas in machine learning.
Many algorithms are based on the practical assumption that training examples are provided in groups, and that a test distribution is represented as a mixture of these groups.
Existing approaches can be categorized into the following three main groups.

\vspace{-2mm}
\paragraph{Sample reweighting}
The most popular approaches involve assigning different training weights to each sample to promote  minority groups, with the weights determined by either group frequency or loss.
Group DRO~\cite{GroupDRO} minimizes the worst-group loss by reweighting samples based on the average loss per group.
Although Group DRO achieves robust results against group distribution shifts, it requires training examples with group supervision.
To handle this limitation, several unsupervised approaches have been proposed.
George~\cite{sohoni2020no} and BPA~\cite{seo2021unsupervised} extend Group DRO to an unsupervised setting by initially training an ERM model and subsequently inferring pseudo-groups through feature clustering.
CVaR DRO~\cite{levy2020large} minimizes the worst loss over all $\alpha$-sized subpopulations, effectively providing an upper bound on the worst-group loss for unknown groups.
LfF~\cite{LfF} simultaneously trains two models, one is with generalized cross-entropy and the other is with the standard cross-entropy loss, and reweights the examples based on their relative difficulty score.
JTT~\cite{JTT} conducts a two-stage procedure, which upweights the examples that are misclassified by the first-stage model.
Idrissi~\etal~\cite{idrissi2022simple} analyze simple data subsampling and reweighting baselines based on group or class frequency to handle dataset imbalance issues.
LWBC~\cite{kim2022learning} employs an auxiliary module to identify bias-conflicted data and assigns large weights to them.

\vspace{-2mm}
\paragraph{Representation learning}
Some approaches aim to learn debiased representations to mitigate spurious correlations directly.
ReBias~\cite{ReBias} employs the Hilbert-Schmidt independence criterion~\cite{gretton2005measuring} to ensure feature representations remain independent of predefined biased representations.
Cobias~\cite{seo2022information} measures bias through conditional mutual information between feature representations and group labels and incorporates this metric as a debiasing regularizer.
IRM~\cite{IRM} learns invariant representations across diverse environments, where the environment variable is treated as equivalent to the group.
While IRM requires supervision for the environment variable, unsupervised alternatives such as EIIL~\cite{creager2021environment} and PGI~\cite{ahmed2020systematic} infer environments by assigning each training example to groups that violate the IRM objective.

\vspace{-2mm}
\paragraph{Post-processing}
While most existing approaches focus on in-processing techniques, such as feature representation learning or sample reweighting during training to improve group robustness, our framework stands apart by addressing group robust optimization through a simple post-processing method based on class-specific score scaling, which requires no additional training.
Although post-processing techniques like temperature scaling~\cite{guo2017calibration} or Platt scaling~\cite{john2000platt} are popular in confidence calibration, they are unsuitable for our task since they scale prediction scores uniformly across classes and do not alter label predictions.
Recently, post-hoc methods have been proposed to retrain the model?s last layer using a group-balanced dataset~\cite{kirichenko2022last} or adjust the final logits~\cite{liu2022avoiding}, but these approaches still involve additional training, differentiating them from our framework.

\section{Proposed Approach}
\label{sec:method}

This section first presents our class-specific scaling technique, which captures the trade-off landscape and identifies the optimal performance points for desired objectives along the trade-off curve.
We also propose an instance-wise class-specific scaling approach to overcome the trade-off and further improve the performance.
Based on the proposed scaling strategy, we introduce a novel and intuitive measure for evaluating the group robustness of an algorithm with consideration of the trade-off.

\subsection{Problem Setup}
\label{sec:setup}
Consider a triplet $(x, y, a)$ with an input feature $x \in \mathcal{X}$, a target label $y \in \mathcal{Y}$, and an attribute $a \in \mathcal{A}$. 
We define groups based on the pair of a target label and an attribute, such that $g := (y, a) \in \mathcal{Y} \times \mathcal{A} =: \mathcal{G}$.
Suppose that the training set consists of $n$ examples without attribute annotations, \eg, $\{(x_1, y_1), ..., (x_n, y_n)\}$, while the validation set includes {$m$ examples with group annotations, \eg, $\{(x_1, y_1, a_1), ..., (x_m, y_m, a_m)\}$, for selecting scaling parameters.
This assumption is known to be essential for model selection or hyperparameter tuning~\cite{GroupDRO, JTT, LfF, idrissi2022simple} although not desirable for the practicality of algorithms. 
However, we will show that our algorithm works well with only a few examples with attribute annotations in the validation set; considering such marginal labeling cost, our approach is a meaningful step to deal with notorious bias problems in datasets and models.
 
Our goal is to learn a model $f_\theta(\cdot): \mathcal{X} \rightarrow \mathcal{Y}$ that is robust to group distribution shifts.
To measure the group robustness, we typically refer to the robust accuracy such as unbiased accuracy (UA) and worst-group accuracy (WA).
The definitions of UA and WA require the group-wise accuracy (GA), which is formally given by
\begin{align}
    \text{GA}_{(r)} := \frac{\sum_{i} \mathds{1}(f_\theta(\mathbf{x}_i) = y_i, g_i=r)}{\sum_{i} \mathds{1}(g_i=r)},
\end{align}
where $\mathds{1}(\cdot)$ denotes an indicator function and $\text{GA}_{(r)}$ is the accuracy of the $r^\text{th}$ group samples.
Then, the robust accuracies are defined by
\begin{align}
    \text{UA} := \frac{1}{|\mathcal{G}|}\sum_{r\in\mathcal{G}}{\text{GA}_{(r)}}~~~\text{and}~~~ \text{WA} := \min_{r\in\mathcal{G}} \text{GA}_{(r)}.
\end{align}
The goal of the group robust optimization is to ensure robust performance in terms of UA or WA regardless of the group membership of a sample.

\begin{figure}[t]
\centering
    \begin{subfigure}[m]{0.85\linewidth}
    	\includegraphics[width=\linewidth]{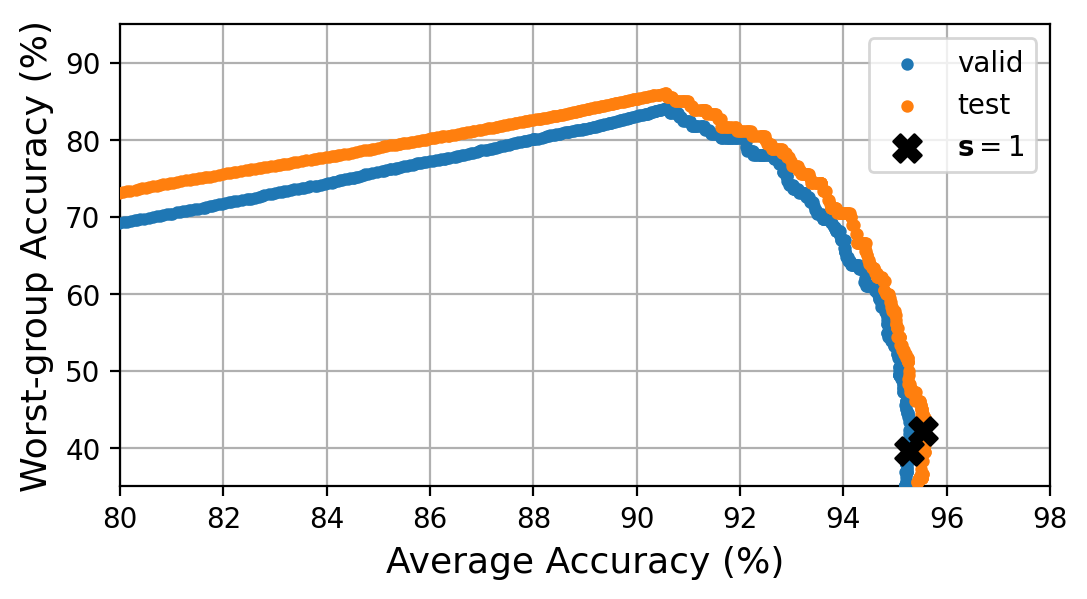}\vspace{-1mm}
	\subcaption{Worst-group accuracy}
	\label{fig:tsne_noise}
	    \vspace{2mm}
    \end{subfigure}
        \begin{subfigure}[m]{0.85\linewidth}
    	\includegraphics[width=\linewidth]{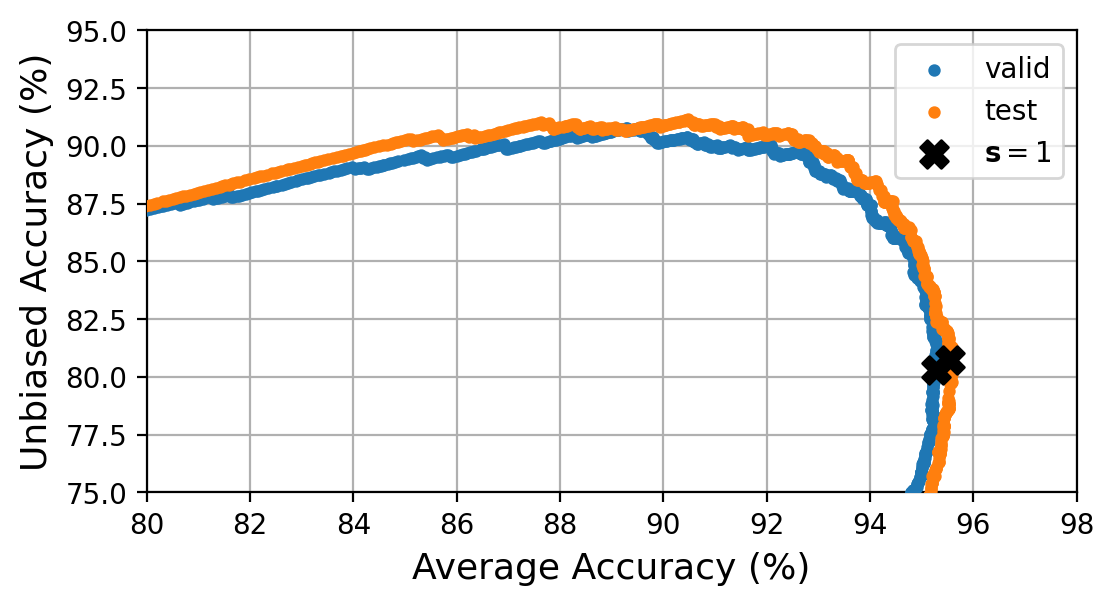}\vspace{-1mm}
	\subcaption{Unbiased accuracy}
	\label{fig:tsne_erm}
	\end{subfigure} 
    \caption{The relation between the robust and average accuracies obtained by varying the class-specific scaling factor $\mathbf{s}$ with ERM on CelebA.
    The black marker denotes the original point, where the uniform scaling is applied. 
    }
    \label{fig:observation_curve}
\end{figure}

 \subsection{Class-Specific Robust Scaling}
 \label{sec:robust_scaling}
 
As illustrated in Figure~\ref{fig:observation_tradeoff}, all algorithms exhibit a clear trade-off between robust accuracy and average accuracy. 
To analyze this behavior more closely, we propose a simple non-uniform scaling method for adjusting the scores associated with individual classes. 
This approach can influence the final decision of the classifier: by upweighting the prediction scores for minority classes, a sample may be classified into a minority class even if its initial score is low. 
Consequently, this adjustment can enhance the worst-group accuracy at the cost of a slight reduction in average accuracy, yielding a favorable trade-off for achieving group robustness.
Formally, the prediction with the class-specific scaling is given by
  \begin{align}
     \argmax_c~(\mathbf{s}\odot \hat{\mathbf{y}})_c,
     \label{eq:rs}
 \end{align}
where $\mathbf{\hat{y}} \in \mathbb{R}^C$ is a prediction score vector over $C$ classes, $\mathbf{s} \in \mathbb{R}^C$ is a $C$-dimensional scaling coefficient vector, and $\odot$ denotes the element-wise product operator.

Based on the ERM model, we obtain a set of the average and robust accuracy pairs using a wide range of the class-specific scaling factors and illustrate their relations in Figure~\ref{fig:observation_curve}.
The black markers indicate the point with a uniform scaling, \ie, $\mathbf{s} = (1, \dots, 1) \in \mathbb{R}^C$.
The graphs show that a simple class-specific scaling effectively represents the landscape of the trade-off between the two accuracies.
This validates the ability to identify the desired Pareto optimal points between the robust and average accuracies in the test set by following two simple steps: 1) finding the optimal class-specific scaling factors that maximize the target objective (UA, WA, or AA) in the validation set, and 2) apply the scaling factors to the test set\footnote{Refer to our supplementary document for the coherency of robust scaling in the validation and test sets.}.
We refer to this scaling strategy for robust prediction as \textit{robust scaling}.
 
To identify the optimal scaling factor $\mathbf{s}$, we perform a greedy search, where we first identify the best scaling factor for a class and then determine the optimal factors of the remaining ones sequentially conditioned on the previously estimated scaling factors.
The greedy search is sufficient for finding good scaling factors partly because there are many different near-optimal solutions.
Thanks to the simplicity of the process, the entire procedure takes negligible time even in large-scale datasets with multiple classes.
It is worth noting that, as a post-processing method, robust scaling can be seamlessly applied to existing robust optimization techniques without requiring additional training. 
Our approach enables the identification of any desired performance point on the trade-off envelope using a pretrained model.
For example, even when dealing with multiple tasks, our robust scaling approach is flexible enough to handle the situation; we only need to apply a scaling factor optimized for each target objective, leaving the trained model unchanged.
Meanwhile, existing robust optimization methods have limited flexibility and require to training separate models for each target objective.

\begin{table*}[t]
\begin{center}
\caption{Experimental results of the robust scaling (RS) and instance-wise robust scaling (IRS) on the CelebA dataset using ResNet-18 with the average of three runs (standard deviations in parenthesis), where RS and IRS are applied to maximize each target metric independently.
\textit{Gain} indicates the average (standard deviations) of performance improvement for each run.
On top of all existing approaches, RS can maximize all target metrics consistently and IRS further boosts the performance.
}
\label{tab:celebA}
 \scalebox{0.8}{
 \hspace{-0.3cm}
\setlength\tabcolsep{6pt} 
\begin{tabular}{lc|cc|cccccc}
\toprule
& Group  & \multicolumn{2}{c|}{Robust Coverage}  & \multicolumn{6}{c}{Accuracy (\%)}\\
Method & Supervision & Worst-group & Unbiased &  Worst-group & (Gain) & Unbiased & (Gain) &  Average  & (Gain) \\
\hline
ERM & & -& -& 34.5 (6.1) & - &77.7 (1.8) &- &{95.5 (0.4)} & - \\
ERM + RS & & {83.0 (0.8)} & {88.1 (0.6)} &  {82.8 (3.3)} & +47.7 (7.8)&  {91.2 (0.5)} &+13.3 (2.0) & \textbf{95.8 (0.2)}& +0.4 (0.2) \\
ERM + IRS  &   &\textbf{83.4 (0.1)} & \textbf{88.4 (0.4)} & \textbf{87.2 (2.0)} & +52.7 (3.3) & \textbf{91.7 (0.2)} &+13.8 (1.6) &\textbf{95.8 (0.1)} & +0.4 (0.3)\\
\hline
CR & & - & -&  70.6 (6.0) & & 88.7 (1.2) & & {94.2 (0.7)}  & \\
CR + RS & &{82.9 (0.5)}  & {88.2 (0.3)}&  {82.7 (5.2)} &+12.2 (7.5)& {91.0 (1.0)} &+2.2 (1.3) & {95.4 (0.5)} & +1.3 (0.4)\\
CR + IRS & & \textbf{83.6 (1.1)} &\textbf{88.6 (0.5)} &\textbf{84.8 (1.5)} & +14.2 (5.2) & \textbf{91.3 (0.4)} & +2.5 (1.4) &\textbf{95.5 (0.1)} & +1.3 (0.3)\\
\hline
SUBY~\cite{idrissi2022simple} & & - & -&65.7 (3.9) &- &87.5 (0.9) &- &{94.5 (0.7)} &- \\
SUBY + RS & & 81.5 (1.0) & 87.4 (0.1) & {80.8 (2.9)} &+15.1 (3.0) & {90.5 (0.8)}  &+3.0 (0.9)  & {95.3 (0.6)} & +0.8 (0.6)\\
SUBY + IRS & & \textbf{82.3 (1.1)} &\textbf{87.8 (0.2)} &\textbf{82.3 (2.0)} & +16.5 (4.1) & \textbf{90.8 (0.8)} & +3.3 (1.1) & \textbf{95.5 (0.3)}  & +1.1 (0.4) \\
\hline
LfF~\cite{LfF} & &-&- &  55.6 (6.6) &- & 81.5 (2.8) &- & {92.4 (0.8)} &- \\
LfF + RS & &{74.1 (3.5)}  &{79.7 (2.6)} & {78.7 (4.1)} &+23.2 (2.5) & {85.4 (2.4)} &+4.0 (0.8) & \textbf{93.4 (0.7)} & +1.0 (0.2) \\
LfF + IRS & &\textbf{74.6 (4.1)}  &\textbf{79.8 (3.1)} & \textbf{78.9 (5.3)} &+23.4 (4.1) & \textbf{86.0 (2.2)} &+4.6 (1.5) & {93.1 (1.5)} & +0.7 (0.5) \\
\hline
JTT~\cite{JTT}  & & - & -& 75.1 (3.6) &- &85.9 (1.4) &- &{89.8 (0.8)} &- \\
JTT + RS  & &{77.3 (0.7)} &{81.9 (0.7)}  & {82.9 (2.3)} &+7.8 (3.0)  &{87.6 (0.5)} &+1.7 (0.4)  &{90.3 (1.3)} & +0.6 (0.1) \\
JTT + IRS  & &\textbf{78.9 (2.1)} &\textbf{82.1 (1.5)}  & \textbf{84.9 (4.5)} &+9.8 (3.7)  &\textbf{88.5 (0.8)} &+2.5 (0.8)  &\textbf{91.0 (1.8)} & +1.2 (0.5) \\
\hline
 GR  & \multirow{3}{*}{\checkmark} & - & -  & 88.6 (1.9) &- &92.0 (0.4) &- &{92.9 (0.8)} &- \\
GR + RS & &{86.9 (0.4)} &{88.4 (0.2)} & {90.0 (1.6)} &+1.4 (1.1) & {92.4 (0.5)} &+0.5 (0.4) & {93.8 (0.4)} & +0.8 (0.5)\\
GR + IRS & & \textbf{87.0 (0.2)} & \textbf{88.6 (0.2)} & \textbf{90.0 (2.3)} & +1.4 (1.8) & \textbf{92.6 (0.6)} & +0.6 (0.4) & \textbf{94.2 (0.3)} & +1.3 (1.0) \\
\hline
SUBG~\cite{idrissi2022simple}  & \multirow{3}{*}{\checkmark} & - & -  - &87.8 (1.2) &- &90.4 (1.2) &- &{91.9 (0.3)} &- \\ 
SUBG + RS & & 83.6 (1.6) & 87.5 (0.7) & {88.3 (0.7)} &+0.5 (0.4) & {90.9 (0.5)} & +0.5 (0.5)  & {93.9 (0.2)} & +1.9 (0.6)\\
SUBG + IRS & & \textbf{84.5 (0.8)} &\textbf{87.9 (0.1)} &\textbf{88.7 (0.6)} & +0.8 (0.7) & \textbf{91.0 (0.3)} & +0.6 (0.9) & \textbf{94.0 (0.2)} & +2.1 (1.0) \\
\hline
Group DRO~\cite{GroupDRO} &\multirow{3}{*}{\checkmark}  & -& - &  88.4 (2.3) &- & 92.0 (0.4) &- & {93.2 (0.8)} &-  \\
Group DRO + RS & &{87.3 (0.2)} & {88.3 (0.2)}&  {89.7 (1.2)} &+1.4 (1.0) & {92.3 (0.1)} &+0.4 (0.2) & {93.9 (0.3)} & +0.7 (0.5) \\
Group DRO + IRS & & \textbf{87.5 (0.4)} & \textbf{88.4 (0.2)}& \textbf{90.0 (2.3)} & +2.6 (1.8) & \textbf{92.6 (0.6)} & +0.6 (0.4) & \textbf{94.7 (0.3)} &+1.5 (1.1) \\
\bottomrule
\end{tabular}
 }
 \vspace{-1mm}
\end{center}
\end{table*}


 \subsection{Instance-wise Robust Scaling}
 \label{sec:method}
The optimal scaling factor can be adaptively applied to each test example, enabling instance-specific scaling to potentially overcome the trade-off and further improve accuracy. 
Previous approaches~\cite{seo2021unsupervised, sohoni2020no} have demonstrated the ability to identify hidden spurious attributes by clustering in the feature space for debiased representation learning. 
Similarly, we take advantage of feature clustering for adaptive robust scaling; we obtain the optimal class-specific scaling factors based on the cluster membership of each sample.
The overall algorithm of our instance-wise robust scaling (IRS) is outlined as follows.
%
\begin{enumerate}
  \item Perform clustering with the validation dataset on the feature space and store the cluster centroids. \vspace{-1mm}
   \item Find the optimal scaling factor for each cluster. \vspace{-1mm}
  \item Apply the estimated scaling factor to each test example based on its cluster membership.
\end{enumerate}
\vspace{2mm}
In step 1, we use a simple \textit{K}-means clustering algorithm.
Empirically, when $K$ is sufficiently large, \ie, $K > 10$, IRS achieves stable and superior results, compared to the original class-specific scaling.

 \subsection{Robust Coverage}
 \label{sec:robust_coverage}
Although robust scaling identifies a desired performance point on the trade-off curve, it captures only a single point, overlooking the other Pareto-optimal solutions. 
To enable a more comprehensive evaluation of an algorithm, we propose a convenient scalar measure that summarizes the robust-average accuracy trade-off.
Formally, we define the \textit{robust coverage} as
\begin{align}
     \text{(Robust coverage)} &:=  \int_{c=0}^1\max_\mathbf{s}\big\{\text{RA}^\mathbf{s}|\text{AA}^\mathbf{s}\geq c\big\}dc 
      \nonumber \\
     & \hspace{-5mm}\approx \sum_{d=0}^{D-1}\frac{1}{D}\max_\mathbf{s}\big\{\text{RA}^\mathbf{s}|\text{AA}^\mathbf{s}\geq \frac{d}{D}\big\},
     \label{eq:coverage}
\end{align}
where $\text{RA}^\mathbf{s}$ and $\text{AA}^\mathbf{s}$ denote the robust and average accuracies, respectively, and $D = 10^3$ is the number of slices used for discretization.
The robust coverage measures the area under the Pareto frontier of the robust-average accuracy trade-off curve, where the maximum operation in~(\ref{eq:coverage}) identifies the Pareto optimum for each threshold.
Depending on the target objective of robust coverage in~(\ref{eq:rs}), we use either WA or UA as the measure of RA.


\let\contextbf\null
\definecolor{Gray}{gray}{0.9}
\newcolumntype{g}{>{\columncolor{Gray}}c}

\section{Experiments}
\label{sec:exp}

\begin{table*}[t]
\begin{center}
\caption{Experimental results of RS and IRS on the Waterbirds dataset using ResNet-50 with the average of three runs (standard deviations in parenthesis), where RS and IRS are applied to maximize each target metric independently.
}
\label{tab:waterbirds}
 \scalebox{0.8}{
 \hspace{-0.3cm}
\setlength\tabcolsep{6pt} 
\begin{tabular}{lc|cc|cccccc}
\toprule
& Group  & \multicolumn{2}{c|}{Robust Coverage}  & \multicolumn{6}{c}{Accuracy (\%)}\\
Method & Supervision & Worst-group & Unbiased &  Worst-group & (Gain) & Unbiased & (Gain) &  Average  & (Gain) \\
\hline
ERM   & & - & - &76.3 (0.8) & - &89.4 (0.6) & - &{97.2 (0.2)} & - \\
ERM + RS  & &{76.1 (1.4)} & {82.6 (1.3)}  &{81.6 (1.9)} &+5.3 (1.3)  &{89.8 (0.5)} &+0.4 (0.4)  
&{97.5 (0.1)} & +0.4 (0.2) \\
ERM + IRS  & &\textbf{83.4 (1.1)} &\textbf{86.9 (0.4)} &\textbf{89.3 (0.5)} & +13.0 (0.9) &\textbf{92.7 (0.4)} &+3.3 (0.7) &\textbf{97.5 (0.3)} &+0.3 (0.4)\\
\hline
CR   & & - & - &76.1 (0.7) & - &89.1 (0.7) & - &{97.1 (0.3)} & - \\
CR + RS  & & {73.6 (2.3)}&{82.0 (1.5)} &{79.4 (2.4)} &+3.4 (1.8)  &{89.4 (1.0)}  &+0.3 (0.4)  &\textbf{97.5 (0.3)} & +0.4 (0.1) \\
CR + IRS  & & \textbf{84.2 (2.5)} & \textbf{88.3 (1.0)} & \textbf{88.2 (2.7)} & +12.2 (2.1) & \textbf{92.1 (0.7)} & +3.1 (0.1) & 97.4 (0.2) & +0.3 (0.2)\\
\hline
SUBY~\cite{idrissi2022simple}   & & - & - &72.8 (4.1) & - &84.9 (0.4) & -  &{93.8 (1.5)} & - \\
SUBY + RS  & & {72.5 (1.0)}&{81.2 (1.4)} &{75.9 (4.4)} &+3.4 (1.8)  &{86.3 (0.9)}  &+2.3 (0.9)  &{95.5 (0.2)} & +1.7 (1.1)\\
SUBY + IRS  & & \textbf{78.8 (2.7)} & \textbf{85.9 (1.0)}  & \textbf{82.1 (4.0)} & +9.3 (1.1) & \textbf{89.1 (0.9)} & +4.2 (1.0)& \textbf{96.2 (0.6)} &+2.4 (1.4) \\
\hline
GR    & \multirow{3}{*}{\checkmark} &- & -&86.1 (1.3) & - &89.3 (0.9) & - &{95.1 (1.3)} & -  \\
GR + RS     & &{83.7 (0.3)} & {86.8 (0.7)}&\textbf{89.3 (1.3)} &+3.2 (2.0)  &{92.0 (0.7)} &+2.7 (1.3)  & {95.4 (1.3)} & +0.4 (0.2)\\
GR + IRS     & & \textbf{84.8 (1.7)} &\textbf{87.4 (0.4)} &89.1 (0.8) & +3.0 (1.6)  & \textbf{92.2 (1.0)} & +2.9 (1.6) & \textbf{95.6 (0.8)} &+0.6 (0.3) \\
\hline
SUBG~\cite{idrissi2022simple}  & \multirow{3}{*}{\checkmark} & - & -  & 86.5 (0.9) & - & 88.2 (1.2) & - & 87.3 (1.1) & - \\
SUBG + RS & & 80.6 (2.0) & 82.3 (2.0) & {87.1 (0.7)} &+0.6 (0.5)  & \textbf{88.5 (1.2)} & +0.3 (0.3)  & {91.3 (0.4)} & +4.0 (0.9)\\
SUBG + IRS & & \textbf{82.2 (0.8)} & \textbf{84.1 (0.8)} & \textbf{87.3 (1.3)} & +0.8 (0.6) & 88.2 (1.2) & +0.0 (0.2) & \textbf{93.5 (0.4)} &+6.2 (1.5)\\
\hline
Group DRO~\cite{GroupDRO}   & \multirow{3}{*}{\checkmark} & - & -&88.0 (1.0) & - &92.5 (0.9) & - &{95.8 (1.8)} & - \\
Group DRO + RS   & & {83.4 (1.1)}&{87.4 (1.4)} &{89.1 (1.7)} &+1.1 (0.8)  &{92.7 (0.8)}  &+0.2 (0.1)  & {96.4 (1.5)} & +0.5 (0.5)\\
Group DRO + IRS   & & \textbf{86.3 (2.3)} & \textbf{90.1 (2.6)} & \textbf{90.8 (1.3)} & +2.8 (1.5) & \textbf{93.9 (0.2)} & +1.4 (0.9) & \textbf{97.1 (0.4)} & +1.2 (0.8) \\
\bottomrule
\end{tabular}
 }
\end{center}
  \vspace{-3mm}
\end{table*}


\subsection{Experimental Setup}
\label{sec:exp_setup}

\paragraph{Implementation details}
Following prior works, we adopt ResNet-18, ResNet-50~\cite{he2016deep}, and DenseNet-121~\cite{huang2017densely}, pretrained on ImageNet~\cite{deng2009imagenet}, as our backbone networks for the CelebA, Waterbirds, and FMoW-WILDS datasets, respectively.
For the text classification dataset, CivilComments-WILDS, we use DistillBert~\cite{sanh2019distilbert}.
We employ the standard $K$-means clustering for IRS, where the number of clusters is set to 20, \ie, $K=20$, for all experiments.
We select the final model with the scaling factor that gives the best unbiased coverage in the validation split.
Our implementations are based on the Pytorch~\cite{paszke2019pytorch} framework and all experiments are conducted on a single NVIDIA Titan XP GPU.
Please refer to our supplementary file for the details about the dataset usage.

\vspace{-2mm}
\paragraph{Evaluation metrics}
We evaluate all algorithms in terms of the proposed unbiased and worst-group coverages for comprehensive evaluation, and additionally use the average, unbiased, and worst-group accuracies for comparisons. 
Following previous works~\cite{GroupDRO, JTT}, we report the adjusted average accuracy instead of the na\"ive version for the Waterbirds dataset due to its dataset imbalance issue; we first calculate the accuracy for each group and then report the weighted average, where the weights are given by the relative portion of each group in the training set.
We ran the experiments three times for each algorithm and report their average and standard deviation.

  \begin{table*}[t]
\begin{center}
\caption{Experimental results on the CivilComments-WILDS dataset using a DistilBert architecture with the average of 3 runs.
}
\label{tab:civilcomments}
 \scalebox{0.8}{
 \hspace{-0.3cm}
\setlength\tabcolsep{6pt} 
\begin{tabular}{lc|cc|cccccc}
\toprule
& Group  & \multicolumn{2}{c|}{Robust Coverage}  & \multicolumn{6}{c}{Accuracy (\%)}\\
Method & Supervision & Worst-group & Unbiased &  Worst-group & (Gain) & Unbiased & (Gain) &  Average  & (Gain) \\
\hline
ERM & & - & - & 54.5 (6.8) & - &75.0 (1.2)  & - & 92.3 (0.4) &-  \\
 ERM + RS  & &57.2 (5.1)  & 70.9 (1.5) & 65.5 (1.2) & +11.0 (2.5) & 78.6 (1.5) & +3.7 (2.4) & \textbf{92.5 (0.3)} &+0.2 (0.1)    \\
 ERM + IRS & & \textbf{59.2 (5.3)} & \textbf{71.2 (2.3)} & \textbf{67.0 (2.3)} & +12.5 (2.7) & \textbf{78.8 (1.1)} & +3.8 (1.7) & \textbf{92.5 (0.3)} & +0.2 (0.1)  \\
 \cdashline{1-10}
  GR   & \multirow{3}{*}{\checkmark} & - & - & 64.7 (1.1) & - & 78.4 (0.2) & - & 87.2 (1.0) & -  \\
 GR + RS  & & 59.0 (2.8) & 69.8 (1.0) & 66.0 (0.5) & +1.3 (0.6) & 78.5 (0.1) & +0.1 (0.1)  & 87.9 (0.8) & +0.7 (0.3) \\
 GR + IRS & & \textbf{59.7 (1.6)} & \textbf{70.1 (0.7)} & \textbf{66.2 (0.4)}& +1.6 (0.7)  & \textbf{78.6 (0.1)}& +0.2 (0.2) & \textbf{88.4 (0.6)} & +1.2 (0.6) \\
 \cdashline{1-10}
 Group DRO  & \multirow{3}{*}{\checkmark} & - & - &67.7 (0.6) & - & 78.4 (0.6) & - & {90.0 (0.1)} & -  \\
 Group DRO + RS  & & 60.6 (0.6) & 71.5 (0.3) & 68.8 (0.7) & +1.1 (0.5) & \textbf{78.8 (0.4)} & +0.4 (0.3) & 90.5 (0.2) & +0.5 (0.3) \\
 Group DRO + IRS & & \textbf{62.1 (0.7)} & \textbf{71.9 (0.2)} & \textbf{69.6 (0.4)} & +1.9 (0.6) & \textbf{78.8 (0.5)} & +0.4 (0.6)  & \textbf{90.8 (0.3)} & +0.8 (0.3)  \\
\bottomrule
\end{tabular}
}
\end{center}
\vspace{-2mm}
\end{table*}

 \begin{table*}[t]
\begin{center}
\caption{Experimental results on the FMoW-WILDS dataset using a DenseNet-121 architecture with the average of 3 runs.
}
\label{tab:fmow}
 \scalebox{0.8}{
 \hspace{-0.3cm}
\setlength\tabcolsep{6pt} 
\begin{tabular}{lc|cc|cccccc}
\toprule
& Group  & \multicolumn{2}{c|}{Robust Coverage}  & \multicolumn{6}{c}{Accuracy (\%)}\\
Method & Supervision & Worst-group & Unbiased &  Worst-group & (Gain) & Unbiased & (Gain) &  Average  & (Gain) \\
\hline
ERM & & - & - & 34.5 (1.4) & - & 51.7 (0.5) & - & 52.6 (0.8) & -\\
 ERM + RS  & & 32.9 (0.4) & 39.4 (1.3) & 35.7 (1.6) & +1.2 (0.4) & 52.3 (0.3)& +0.6 (0.3) & 53.1 (0.8) & +0.6 (0.3) \\
 ERM + IRS  & & \textbf{35.1 (0.2)} & \textbf{40.2 (1.1)} & \textbf{36.2 (1.4)} & +1.7 (0.3) & \textbf{52.4 (0.2)} & +0.7 (0.4) &  \textbf{53.4 (0.9)} & +0.8 (0.4)  \\
 \cdashline{1-10}
 GR   & \multirow{3}{*}{\checkmark} & - & - & 31.4 (1.1)	 & - & 49.0 (0.9) & - & 50.1 (1.3)& - \\
 GR + RS  & & 30.2 (1.2) & 37.7 (0.6) & 35.5 (0.4) & +4.2 (0.7) & 49.8 (0.7) & +0.8 (0.3) & 50.7 (1.2) & +0.6 (0.1)  \\
 GR + IRS & & \textbf{31.7 (1.0)} & \textbf{38.9 (2.1)} & \textbf{35.7 (0.9)} & +4.4 (0.4) & \textbf{50.1 (0.6)} & +1.1 (0.3) & \textbf{50.8 (1.4)}& +0.7 (0.1) 	\\
 \cdashline{1-10}
 Group DRO  & \multirow{3}{*}{\checkmark} & - & - & 33.7 (2.0) & - & 50.4 (0.7) & - & 52.0 (0.4) & -  \\
 Group DRO + RS  & & 30.8 (1.8) & 38.2 (0.7) & 36.0 (2.4)	 & +2.3 (0.4) & 50.9 (0.6) & +0.4 (0.4) & 52.4 (0.2) & +0.5 (0.2)\\
 Group DRO + IRS & & \textbf{34.1 (0.8)} & \textbf{40.7 (0.5)} & \textbf{36.4 (2.3)}& +2.7 (0.4)  & \textbf{51.1 (0.3)} & +0.7 (0.5) & \textbf{52.7 (0.2)} & +0.7 (0.2) \\
\bottomrule
\end{tabular}
}
\vspace{-4mm}
\end{center}

\end{table*}

\subsection{Results}

\paragraph {CelebA}
Table~\ref{tab:celebA} presents the experimental results of our robust scaling methods (RS and IRS) on top of the existing approaches including CR, SUBY, LfF, JTT, Group DRO$^\ast$, GR$^\ast$, and SUBG$^\ast$\footnote{A brief introduction to these methods is provided in the supplementary document.}} on the CelebA dataset, where `$\ast$' indicates the method that requires the group supervision in training sets.
In this evaluation, RS and IRS choose scaling factors to maximize individual target metrics---worst-group, unbiased, and average accuracies\footnote{Since our robust scaling strategy is a simple post-processing method, we do not need to retrain models for each target measure and the cost is negligible, taking only a few seconds for each target metric.}.
As shown in the table, our robust scaling strategies consistently improve the performance for all target metrics.
In terms of the robust coverage and robust accuracy after scaling, LfF and JTT are not superior to ERM on the CelebA dataset although their robust accuracies without scaling are much higher than ERM.
The methods that leverage group supervision such as Group DRO and GR achieve better robust coverage results than the others, which verifies that group supervision helps to improve overall performance.
For the group-supervised methods, our scaling technique achieves relatively small performance gains in robust accuracy since the gaps between robust and average accuracies are small and the original results are already close to the optimal robust accuracy.
Note that, compared to RS, IRS further boosts the robust coverage and all types of accuracies consistently in all algorithms.

\vspace{-2mm}
\paragraph{Waterbirds}
Table~\ref{tab:waterbirds} demonstrates the outstanding performance of our approaches with all baselines on the Waterbirds dataset.
Among the compared algorithms, GR and SUBG are reweighting and subsampling methods based on group frequency, respectively.
Although the two baseline approaches exhibit competitive robust accuracy, the average accuracy of SUBG is far below than GR (87.3\% vs. 95.1\%).
This is mainly because SUBG drops a large portion of training samples ($95\%$) to make all groups have the same size, resulting in the significant loss of average accuracy.
Subsampling generally helps to achieve high robust accuracy, but it degrades the overall trade-off as well as the average accuracy, consequently hindering the benefits of robust scaling.
{This observation is coherent to our main claim; the optimization towards the robust accuracy is incomplete and more comprehensive evaluation criteria are required to understand the exact behavior of debiasing algorithms.}
Note that GR outperforms SUBG in terms of all accuracies after adopting the proposed RS or IRS.

\vspace{-2mm}
\paragraph{CivilComments-WILDS}

We also validate the effectiveness of the proposed approach in a large-scale text classification dataset, CivilComments-WILDS~\cite{koh2021wilds}, which has 8 attribute groups.
As shown in Table~\ref{tab:civilcomments}, our robust scaling strategies still achieve meaningful performance improvements for all baselines on this dataset.
Although group-supervised baselines such as GR and Group DRO accomplish higher robust accuracies than the ERM without scaling, ERM benefits from RS and IRS greatly.
ERM+IRS outperforms both Group DRO and GR in average accuracy while achieving competitive worst-group and unbiased accuracies, even without group supervision in training samples and extra training.

\vspace{-2mm}
\paragraph{FMoW-WILDS}
FMoW-WILDS~\cite{koh2021wilds} is a high-resolution satellite imagery dataset with 65 classes and 5 attribute groups, which involves domain shift issues as train, validation, and test splits come from different years.
We report the results from our experiments in Table~\ref{tab:fmow}, which shows that GR and Group DRO have inferior performance even compared with ERM.
On the other hand, our robust scaling methods do not suffer from any performance degradation and  even enhance all kinds of accuracies substantially.
This fact supports the strengths and robustness of our framework in more challenging datasets with distribution shifts.


\subsection{Analysis}

\paragraph{Validation set sizes}
We analyze the impact of the validation set size on the robustness of our algorithm.
Table~\ref{tab:val_size} presents the ERM results on the CelebA dataset by varying the validation set size to $\{100\%, 50\%, 10\%, 1\%\}$ of its full size.
Note that other approaches also require validation sets with group annotations for early stopping and hyperparameter tuning, which are essential to achieve high robust accuracy.
As shown in the table, with only $10\%$ or $50\%$ of the validation set, both RS and IRS achieve almost equivalent performance to the versions with the entire validation set.
Surprisingly, even only $1\%$ of the validation set is enough for RS to gain sufficiently high robust accuracy but inevitably entails a large variance of results.
On the other hand, IRS suffers from performance degradation when only $1\%$ of the validation set is available.
This is mainly because IRS takes advantage of feature clustering on the validation set, which would need more examples for stable results.
In overall, our robust scaling strategies generally improve performance substantially even with a limited number of validation examples with group annotations for all cases.

\begin{table}[t]
\begin{center}
\caption{Ablation study on the size of validation set in our robust scaling strategies on CelebA.
}
\vspace{-1mm}
\label{tab:val_size}
 \scalebox{0.8}{
\hspace{-0.2cm}
\setlength\tabcolsep{6pt} 
\begin{tabular}{lc|cccc}
\toprule
Method & Valid set size & Worst-group & Gain & Unbiased & Gain \\
\hline
ERM & - & 34.5 (6.1) &- & 77.7 (1.8) &- \\
\hline
+ RS & 100\%  & 82.8 (3.3) & \textbf{+48.3} & 91.2 (0.5) & \textbf{+13.5} \\
+ RS & \ \ 50\%  & 83.3 (3.7) & \textbf{+48.8}  & 91.5 (0.9) & \textbf{+13.8} \\
+ RS & \ \ 10\%  & 82.4 (4.3) & \textbf{+48.0} & 91.4 (0.8) & \textbf{+13.7}  \\
+ RS & \ \ \ \ 1\% & 79.2 (10.3) & \textbf{+44.7}  & 90.8 (2.2) & \textbf{+13.1} \\
\hline
+ IRS & 100\%  & 88.7 (0.9) & \textbf{+54.2}  & 92.0 (0.3) & \textbf{+14.3}  \\
+ IRS & \ \ 50\%  & 86.9 (2.0) & \textbf{+52.4} & 91.8 (0.4) & \textbf{+14.1} \\
+ IRS & \ \ 10\%  & 84.4 (6.3) & \textbf{+50.0} & 91.4 (1.0)& \textbf{+13.7}  \\
+ IRS & \ \ \ \ 1\% & 60.4 (14.4) & \textbf{+25.9} & 85.8 (3.2) & \textbf{\ \ +8.0} \\
\bottomrule
\end{tabular}
 }
\vspace{-5mm}
\end{center}
\end{table}

\begin{figure}[t!]
\centering
 \begin{subfigure}[m]{0.85\linewidth}
    	\includegraphics[width=\linewidth]{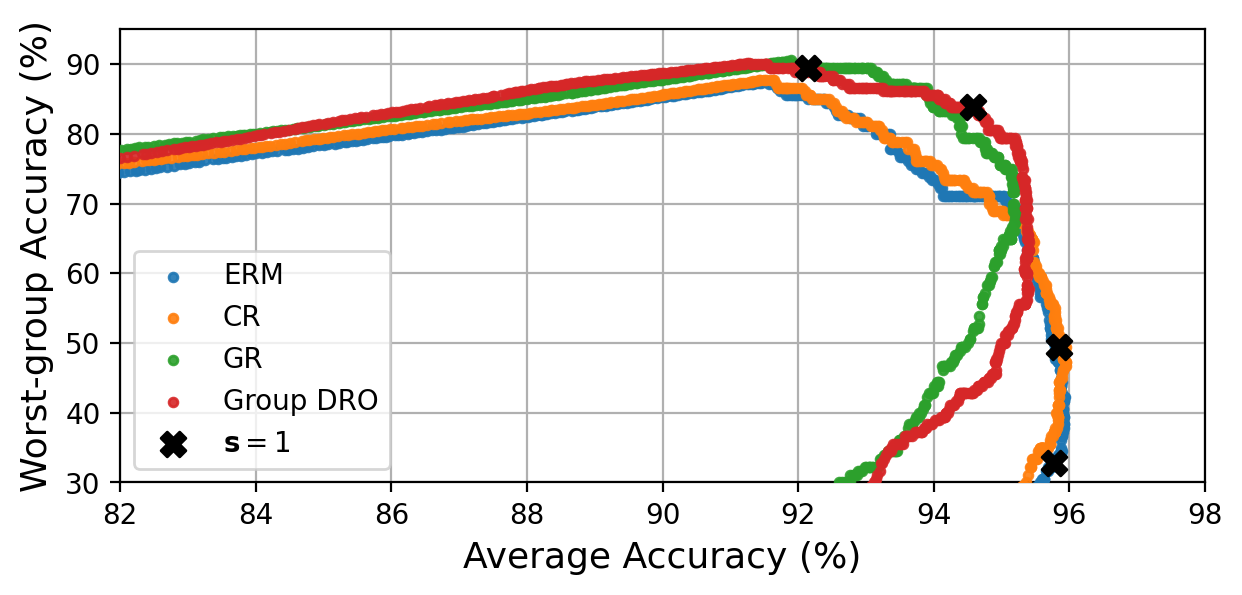}
	\subcaption{With the worst-group accuracy}
	\label{fig:worst_curve_all_celeba}
	\vspace{2mm}
	\end{subfigure} 
	    \begin{subfigure}[m]{0.85\linewidth}
    	\includegraphics[width=\linewidth]{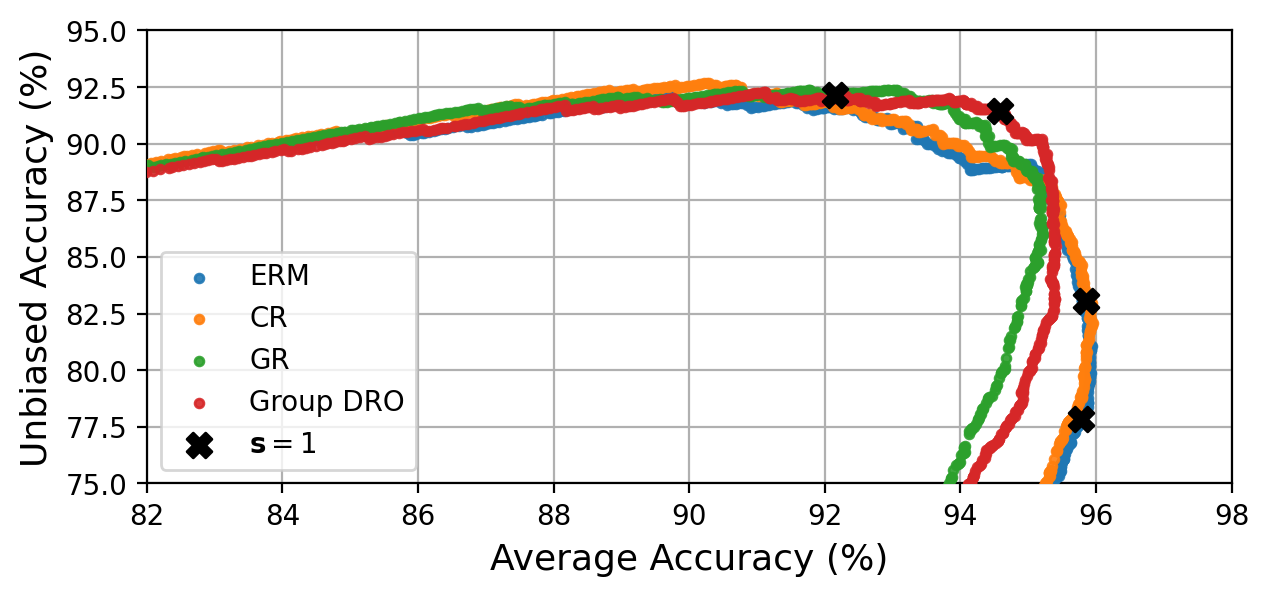}
	\subcaption{With the unbiased accuracy}
	\label{fig:unbias_curve_all_celeba}
	\end{subfigure}
    \caption{
    The robust-average accuracy trade-off curves of various baselines on the CelebA dataset.
        The black marker denotes the original point, where the uniform scaling is applied. 
    }
    \vspace{-1mm}
    \label{fig:robust_all_celeba}
\end{figure}

\vspace{-2mm}
\paragraph{Accuracy trade-off}
Figure~\ref{fig:robust_all_celeba} depicts the robust-average accuracy trade-offs of several existing algorithms on the CelebA dataset.
The black markers denote the points without scaling, implying that there is room for improvement in robust accuracy along the trade-off curve.

\vspace{-2mm}
\paragraph{Number of clusters}

\begin{figure}[t!]
\centering
 \begin{subfigure}[m]{0.85\linewidth}
    	\includegraphics[width=\linewidth]{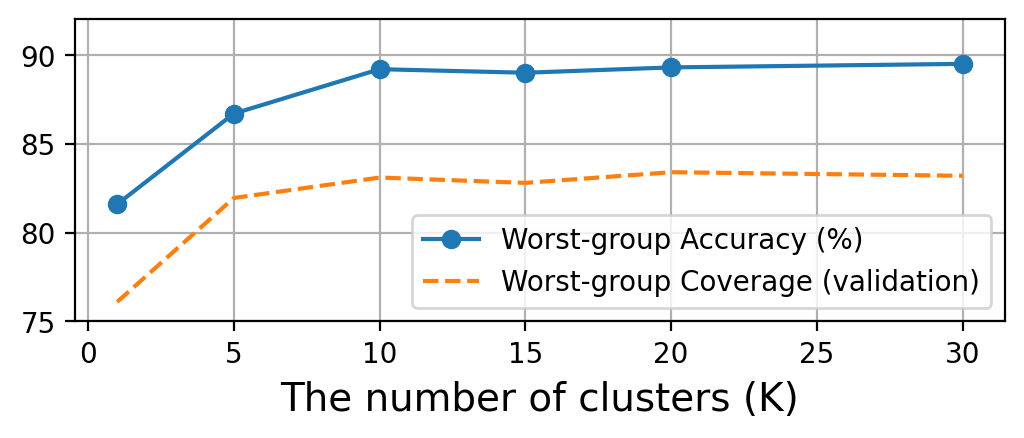}
	\subcaption{With the worst-group accuracy}
	\label{fig:abl_k_worst}
\vspace{2mm}
	\end{subfigure} 
	    \begin{subfigure}[m]{0.85\linewidth}
    	\includegraphics[width=\linewidth]{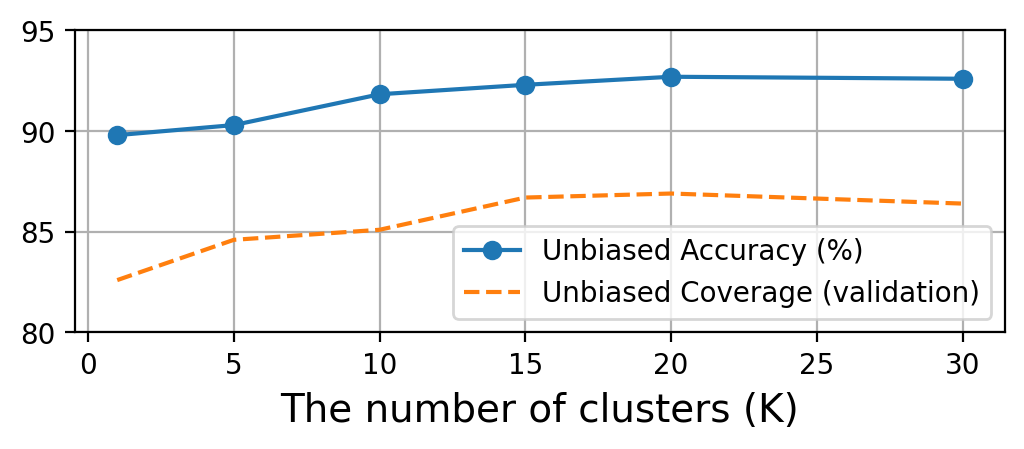}
	\subcaption{With the unbiased accuracy}
	\label{fig:abl_k_unbias}
	\end{subfigure}
    \caption{Sensitivity analysis with respect to the number of clusters in IRS on Waterbirds.
The tendency of the robust coverage in the validation split (orange) is similar with the robust accuracy in the test split (blue).
    }
    \label{fig:abl_k}
\vspace{-3mm}
\end{figure}

We adjust the number of clusters for feature clustering in IRS on the Waterbirds dataset.
Figure~\ref{fig:abl_k} illustrates that the worst-group and unbiased accuracies gradually improve as $K$ increases and are stable with a sufficiently large $K (>10)$.
The leftmost point ($K=1$) denotes RS in each figure.
We also plot the robust coverage results in the validation split, which are almost consistent with the robust accuracy measured in the test dataset.

\vspace{-2mm}
\paragraph{Comparison to reweighting or resampling techniques}
As mentioned in Section~\ref{sec:related}, most existing debiasing techniques~\cite{GroupDRO, JTT, LfF, seo2021unsupervised, idrissi2022simple, kirichenko2022last}, in principle, perform reweighting and/or resampling of training data.
Our approach has a similar idea, but, instead of giving favor to the examples in minority groups during training and boosting their classification scores indirectly via iterative model updates, we directly adjust their classification scores by class-wise scaling after training, thus it gives similar but clearer effects on the results.
As shown in Figure~\ref{fig:robust_all_celeba}, although class reweighting (CR) improves the robust accuracy, this in fact identifies one of the Pareto optimal points on the trade-off curve of ERM obtained by class-specific scaling.
However, because class reweighting employs a single fixed reweighting factor during training based on class frequency, it only reflects a single point and has limited flexibility compared to our wide range of scaling search.
If CR employs a wide range of reweighting factors, then it can identify additional optimal points and achieve additional performance gains, but it requires training separate models for each factor, which is not realistic.
Note that our method can be easily applied to CR or other methods, which allows us to identify more desirable optimal points on the trade-off curve with negligible computational overhead.

\section{Conclusion}
\label{sec:conclusion}
We presented a simple but effective post-processing method that provides a novel perspective of group robustness.
Our work starts from the observation that there exists a clear trade-off between robust and average accuracies in existing works.
From this observation, we first proposed the robust scaling strategy, which captures the full trade-off landscape and identifies any desired performance point on the trade-off curve with no extra training.
Moreover, we proposed an instance-wise robust scaling algorithm that is effective to enhance the trade-off itself.
Based on these strategies, we introduced a novel convenient measure that summarizes the trade-off from a Pareto optimal perspective for a comprehensive evaluation of group robustness.
We believe that our approaches are helpful for analyzing the exact behavior of existing debiasing methods and paving the way in the future research direction.

\vspace{-2mm}
\paragraph{Acknowledgements}
This work was partly supported by Samsung Advanced Institute of Technology (SAIT), and by the Institute of Information communications Technology Planning \& Evaluation (IITP) grants [No.RS-2022-II220959 (No.2022-0-00959), No.RS-2021-II211343, No.RS-2021-II212068], and the National Research Foundation of Korea (NRF) grant [No.RS-2022-NR070855], funded by the Korean government (MSIT).

\bibliographystyle{ieee_fullname}
\bibliography{neurips_2023}


\newpage
\appendix
\onecolumn

\renewcommand{\thesection}{\Alph{section}} 
\renewcommand{\thetable}{A\arabic{table}}
\renewcommand{\thefigure}{A\arabic{figure}}

\clearpage
 
\section{Comparisons} 
\label{sec:comparison}
Below is a brief introduction of the comparisons used in our experiments.

 \paragraph{ERM}
 Given a loss function $\ell(\cdot)$, the objective of empirical risk minimization is optimizing the following loss over training data: 
 \begin{align}
    \min_\theta \Big\{\frac{1}{n}\sum_{i=1}^n \ell(f_\theta(x_i), y_i) \Big\}.
 \end{align}
 
 \paragraph{Class reweighting (CR)}
 To mitigate the class imbalance issue, we can simply reweight the samples based on the inverse of class frequency in the training split, 
  \begin{align}
    \min_\theta \Big\{\frac{1}{n}\sum_{i=1}^n \omega_i \ell(f_\theta(x_i), y_i) \Big\}~~\text{where}~~\omega_i = \frac{n}{\sum_j \mathds{1}(y_j=y_i)}.
 \end{align}
  
 \paragraph{LfF}
Motivated by the observation that bias-aligned samples are more easily learned, LfF~\cite{LfF} simultaneously trains a pair of neural network $(f_B, f_D)$.
The biased model $f_B$ is trained with generalized cross-entropy loss which intends to amplify bias, while the debiased model $f_D$ is trained with a standard cross-entropy loss, where each sample $(x_i, y_i)$ is reweighted by the following relative difficulty score:
\begin{align}
    \omega_i = \frac{\ell(f^B_\theta(x_i), y_i)}{\ell(f^B_\theta(x_i), y_i) + \ell(f^D_\theta(x_i), y_i)}.
\end{align}

 \paragraph{JTT}
 JTT~\cite{JTT} consists of two-stage procedures. In the first stage, JTT trains a standard ERM model $\hat{f}(\cdot)$ for several epochs and identifies an error set $E$ of training examples that are misclassified:
 \begin{align}
E := \{(x_i, y_i)~~\text{s.t.}~\hat{f}(x_i) \neq y_i \}.
 \end{align}
 Next, they train a final model $f_\theta(\cdot)$ by upweighting the examples in the error set $E$ as
 \begin{align}
      \min_\theta \Big\{  \lambda_\text{up}\sum_{(x,y)\in E}\ell(f_\theta(x), y) + \sum_{(x,y)\notin E}\ell(f_\theta(x), y) \Big\}.
 \end{align}

 \paragraph{Group DRO}
Group DRO~\cite{GroupDRO} aims to minimize the empirical worst-group loss formulated as:
 \begin{align}
  \min_\theta \Big\{
     \max_{g\in \mathcal{G}} \frac{1}{n_g} \sum_{i|g_i=g}^{n_g} \ell(f_\theta(x_i), y_i) \Big\}
 \end{align}
 where $n_g$ is the number of samples assigned to $g^\text{th}$ group.
 Unlike previous approaches, group DRO requires group annotations $g=(y,a)$ on the training split.

\paragraph{Group reweighting (GR)}
 Using group annotations, we can extend class reweighting method to group reweighting one based on the inverse of group frequency in the training split, \ie,
   \begin{align}
   & \min_\theta \Big\{\frac{1}{n}\sum_{i=1}^n \omega_i \ell(f_\theta(x_i), y_i) \Big\} \nonumber \\
    &~~~~~~~~~~ \text{where}~~ \omega_i = \frac{n}{\sum_j \mathds{1}(y_j=y_i, a_j=a_i)}
  \end{align}
  
  \paragraph{SUBY/SUBG} To mitigate the data imbalance issue, SUBY subsample majority classes, so all classes have the same size with the smallest class on the training dataset, as in~\cite{idrissi2022simple}.
  Similarly, SUBG subsample majority groups.

%

\section {Attribute-specific Robust Scaling with Group Supervision}
\label{sec:ars}
If the supervision of group (spurious-attribute) information can be utilized during our robust scaling, it will provide flexibility to further improve the performance.
To this end, we first partition the examples based on the values of spurious attributes and find the optimal scaling factors for each partition separately.
Like as the original robust scaling procedure, we obtain the optimal scaling factors for each partition in the validation split and apply them to the test split.
However, this partition-wise scaling is basically unavailable because we do not know the spurious attribute values of the examples in the test split and thus cannot partition them, 
In other words, we need to estimate the spurious-attribute values in the test split for partitioning.
To conduct attribute-specific robust scaling (ARS), we follow a simple algorithm described below:

\begin{enumerate}
\item Partition the examples in the validation split by the values of the spurious attribute.
\item Find the optimal scaling factors for each partition in the validation split. 
\item Train an independent estimator model to classify spurious attribute. 
\item Estimate the spurious attribute values of the examples in the test split using the estimator, and partition the test samples according to their estimated spurious attribute values. 
\item For each sample in the test split, apply the optimal scaling factors obtained in step 2 based on its partition.
\end{enumerate}

To find a set of scale factors corresponding to each partition, we adopt a na\"ive greedy algorithm that performed in one partition at a time.
This attribute-specific robust scaling further increases the robust accuracy compared to the original robust scaling, and also improves the robust coverage, as shown in Table~\ref{tab:grs_all_supple}.
Note that our attribute-specific scaling strategy allows ERM to match the supervised state-of-the-art approach, Group DRO~\cite{GroupDRO}.

One limitation is that it requires the supervision of spurious attribute information to train the estimator model in step 3. 
However, we notice that only a very few examples with the supervision is enough to train the spurious-attribute estimator, because it is much easier to learn as the word ``spurious correlation" suggests.
To determine how much the group-labeled data is needed, we train several spurious-attribute estimators by varying the number of group-labeled examples, and conduct ARS using the estimators.
Table~\ref{tab:abl_groupsize} validates that, compared to the overall training dataset size, a very small amount of group-labeled examples is enough to achieve high robust accuracy.

 \begin{table*}[t]
\begin{center}
\caption{Results of the attribute-specific robust scaling (ARS) on the CelebA and Waterbirds datasets with the average of three runs (standard deviations in parenthesis), where ARS is applied to maximize each target metric independently.
Note that our post-processing strategy, ARS, allows ERM to achieve competitive performance to Group DRO that utilizes the group supervision during training.
}
\label{tab:grs_all_supple}
 \scalebox{0.85}{
 \hspace{-0.3cm}
\setlength\tabcolsep{8pt} 
\begin{tabular}{cl|cc|ccc}
\toprule
 & & \multicolumn{2}{c|}{Robust Coverage} & \multicolumn{3}{c}{Accuracy (\%)} \\
Dataset & Method & Worst. & Unbiased & Worst. & {Unbiased} &  Average \\
\hline
 \multirow{3}{*}{CelebA} & ERM  & - & - & 34.5 (6.1) &77.7 (1.8) &95.5 (0.4) \\
& ERM + ARS  &\textbf{87.6 (1.0)} &\textbf{89.0 (0.2)}&\textbf{88.5 (1.8)} &{91.9 (0.3)}  &\textbf{95.8 (0.1)} \\
\cdashline{2-7}
& Group DRO & 87.3 (0.2) & 88.3 (0.2) & 88.4 (2.3) & \textbf{92.0 (0.4}) & 93.2 (0.8) \\
\hline
 \multirow{3}{*}{Waterbirds}&ERM  & - & - &76.3 (0.8) &89.4 (0.6) &{97.2 (0.2)} \\
& ERM + ARS      &\textbf{84.4 (1.9)} &\textbf{87.8 (1.7)} &\textbf{89.3 (0.4)} &\textbf{92.5 (0.4)} &\textbf{97.5 (1.0)} \\
\cdashline{2-7}
& Group DRO & {83.4 (1.1)}&{87.4 (2.3)} &88.0 (1.0) &\textbf{92.5 (0.9)} &{95.8 (1.8)} \\
\bottomrule
\end{tabular}
 }
\end{center}

\end{table*}


\begin{table}[t]
\begin{center}
\caption{Effects of the size of group-labeled examples on the attribute-specific robust scaling on the CelebA dataset.
Group-labeled size denotes a ratio of group-labeled samples among all training examples for training estimators.
Spurious accuracy indicates the average accuracy of spurious-attribute classification using the estimators on the test split.
}
\vspace{0.3cm}
\label{tab:abl_groupsize}
 \scalebox{0.8}{
\setlength\tabcolsep{7pt} 
\begin{tabular}{c||c|ccc|cc}
\toprule
&Accuracy (\%)& \multicolumn{3}{c|}{Accuracy (\%)} & \multicolumn{2}{c}{Robust Coverage} \\
Group-labeled size & Spurious &Worst-group & Unbiased &  Average & Worst-group & Unbiased \\
\hline
100\% &98.4 &{89.1 (3.0)} &{92.4 (1.1)} &{93.1 (1.2)} & {87.6 (1.0)} &{89.0 (0.5)} \\
10\%  & 97.7 &88.5 (1.8) &91.9 (0.3) &92.8 (0.6) &86.8 (0.4) &89.0 (0.2) \\
1\%   & 95.8 &88.5 (1.8) &91.9 (0.3) &92.9 (0.6) &87.1 (0.3) &89.0 (0.2) \\
0.1\%&92.6  &88.4 (2.1) &91.8 (0.5) &92.4 (0.8) &87.1 (0.3) &89.0 (0.2) \\
\bottomrule
\end{tabular}
 }
\end{center}
\end{table}

\section{Experimental Details}
\label{sec:exp_detail}

\subsection{Datasets}
\label{sec:datasets_detail}
CelebA~\cite{CelebA} is a large-scale dataset for face image recognition, consisting of 202,599 celebrity images, with 40 attributes labeled on each image.
Among the attributes, we primarily examine \textit{hair color} and \textit{gender} attributes as a target and spurious attributes, respectively.
We follow the original train-validation-test split~\cite{CelebA} for all experiments in the paper.
Waterbirds~\cite{GroupDRO} is a synthesized dataset, which are created by combining bird images in the CUB dataset~\cite{wah2011caltech} and background images from the Places dataset~\cite{zhou2017places}, consisting of 4,795 training examples.
The two attributes---one is the type of bird, \{waterbird, landbird\} and the other is background places, \{water, land\}, are used for the experiments with this dataset.
CivilComments-WILDS~\cite{koh2021wilds} is a large-scale text dataset, which has 269,038 training comments, 45,180 validation comments, and 133,782 test comments. 
This task is to classify whether an online comment is toxic or not, which is spuriously correlated to demographic identities (\textit{male, female, White, Black, LGBTQ, Muslim, Christian, and other religion}).
FMoW-WILDS~\cite{koh2021wilds} is based on the Functional Map of the World dataset~\cite{christie2018functional}, comprising high-resolution satellite images from over 200 countries and over the years 2002-2018. 
The label is one of 62 building or land use categories, and the attribute represents both the year and geographical regions (\textit{Africa, the Americas, Oceania, Asia, or Europe}).
It consists of 76,863 training images from the years 2002-2013, 19,915 validation images from the years 2013-2016, and 22,108 test images from the years 2016-2018.

\subsection{Class-specific Scaling} 
To identify the optimal points, we obtain a set of the average and robust accuracy pairs using a wide range of the class-specific scaling factors, \ie, $\mathbf{s}_i= (1.05)^n~\text{for}-200 \leq n \leq 200$ for $i^\text{th}$ class. Note that we search for the scaling factor of each class in a greedy manner, as stated in Section~\ref{sec:robust_scaling}.

\subsection{Hyperparameter Tuning}

We tune the learning rate in $\{10^{-3}, 10^{-4}, 10^{-5}, 10^{-6}\}$ and the weight decay in $\{1.0, 0.1, 10^{-2}, 10^{-4}\}$ for all baselines on all datasets.
We used 0.5 of $q$ for LfF.
For JTT, we searched $\lambda_\text{up}$ in $\{20, 50, 100\}$ and updated the error set every epoch for CelebA dataset and every 60 epochs for Waterbirds dataset.
For Group DRO, we tuned $C$ in $\{0, 1, 2, 3, 4\}$, and used 0.1 of $\eta$.

\begin{table}[t]
\begin{center}
\caption{Realized robust coverage results on the Waterbirds and CelebA datasets with the average of three runs (standard deviations in parenthesis).
}
\vspace{0.3cm}
\label{tab:modified_rc}
 \scalebox{0.8}{
\setlength\tabcolsep{10pt} 
\begin{tabular}{cl|cc|cc}
\toprule
& & \multicolumn{2}{c|}{Robust Coverage} & \multicolumn{2}{c}{Realized Robust Coverage} \\
Dataset & Method & Worst-group & Unbiased & Worst-group & Unbiased \\
\hline
Waterbirds & ERM & 70.3 (1.3) 	& 79.4 (0.7) & 69.0 (1.5) & 78.7 (0.8) \\
Waterbirds & CR & 68.9 (1.1) 		& 78.5 (0.5) & 67.8 (1.2) & 77.9 (0.4) \\
Waterbirds & Group DRO 			& 80.8 (0.6) & 85.2 (0.1) & 78.6 (1.0) & 83.8 (0.4) \\
Waterbirds & GR & 78.8 (5.6) 		& 83.7 (0.7) & 77.9 (1.4) & 82.8 (0.8)\\
 \hline
CelebA & ERM & 78.9 (1.7) 		& 86.0 (0.6) & 75.9 (2.2) & 85.4 (0.7) \\
CelebA &  CR & 77.2 (2.8) 		& 85.6 (0.9) & 71.8 (1.3) & 85.0 (0.6) \\
CelebA &  Group DRO & 84.2 (0.6) 	& 86.7 (0.5) & 81.0 (1.7) & 86.1 (0.2) \\
CelebA &  GR & 84.2 (0.5) 		& 87.5 (0.3) & 81.2 (1.6) & 87.0 (0.5) \\
\bottomrule
\end{tabular}
 }
\end{center}
\end{table}


 \section{Additional Results}

 \begin{figure}[t]
\centering
    \begin{subfigure}[m]{0.4\linewidth}
    	\includegraphics[width=\linewidth]{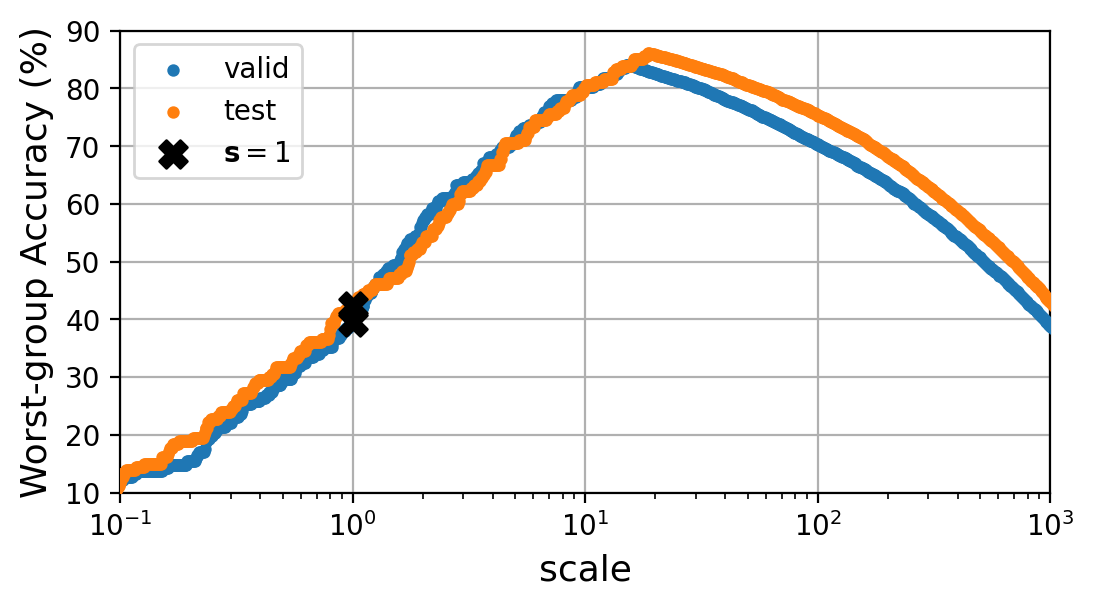}
	\subcaption{Worst-group accuracy}
	\label{fig:tsne_noise}
    \end{subfigure} 
    	\hspace{0.5cm}
        \begin{subfigure}[m]{0.4\linewidth}
    	\includegraphics[width=\linewidth]{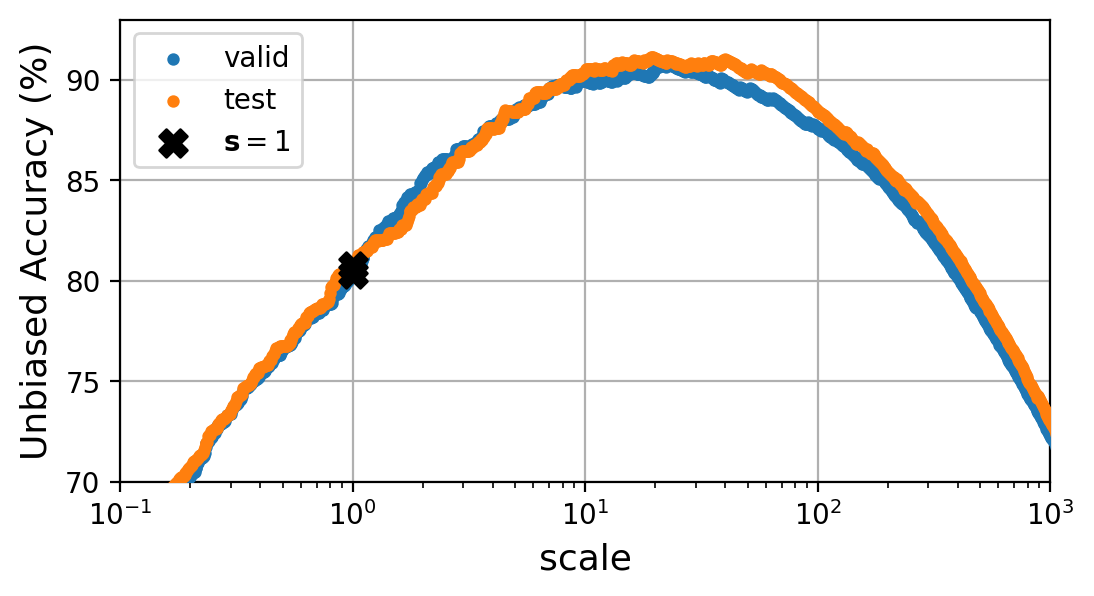}
	\subcaption{Unbiased accuracy}
	\label{fig:tsne_erm}
	\end{subfigure} 
    \caption{Effects of varying the class-specific scaling factors on the robust accuracy using ERM model on the CelebA dataset.
    Since this experiment is based on the binary classifier, a single scaling factor is varied with the other fixed to one. 
These results show that the optimal scaling factor identified in the validation set can be used in the test set to get the final robust prediction.
    }
    \label{fig:observation_scale_curve}
\end{figure}

\paragraph{Feasibility}
\label{sec:analysis}
We visualize the relationship between scaling factors and robust accuracies in Figure~\ref{fig:observation_scale_curve}, where the curves are constructed based on validation and test splits are sufficiently well-aligned to each other.
This implies that the optimal scaling factor identified in the validation set can be used in the test set to get the final robust prediction.

\begin{figure}[t!]
\centering
 \begin{subfigure}[m]{0.47\linewidth}
    	\includegraphics[width=\linewidth]{figures/worst_curve_all.png}
	\subcaption{Worst-group accuracy}
	\label{fig:worst_curve_all_celeba}
	\vspace{0.2cm}
	\end{subfigure} 
	\hspace{0.5cm}
	    \begin{subfigure}[m]{0.47\linewidth}
    	\includegraphics[width=\linewidth]{figures/unbias_curve_all.png}
	\subcaption{Unbiased accuracy}
	\label{fig:unbias_curve_all_celeba}
	\end{subfigure}
	 \begin{subfigure}[m]{0.47\linewidth}
    	\includegraphics[width=\linewidth]{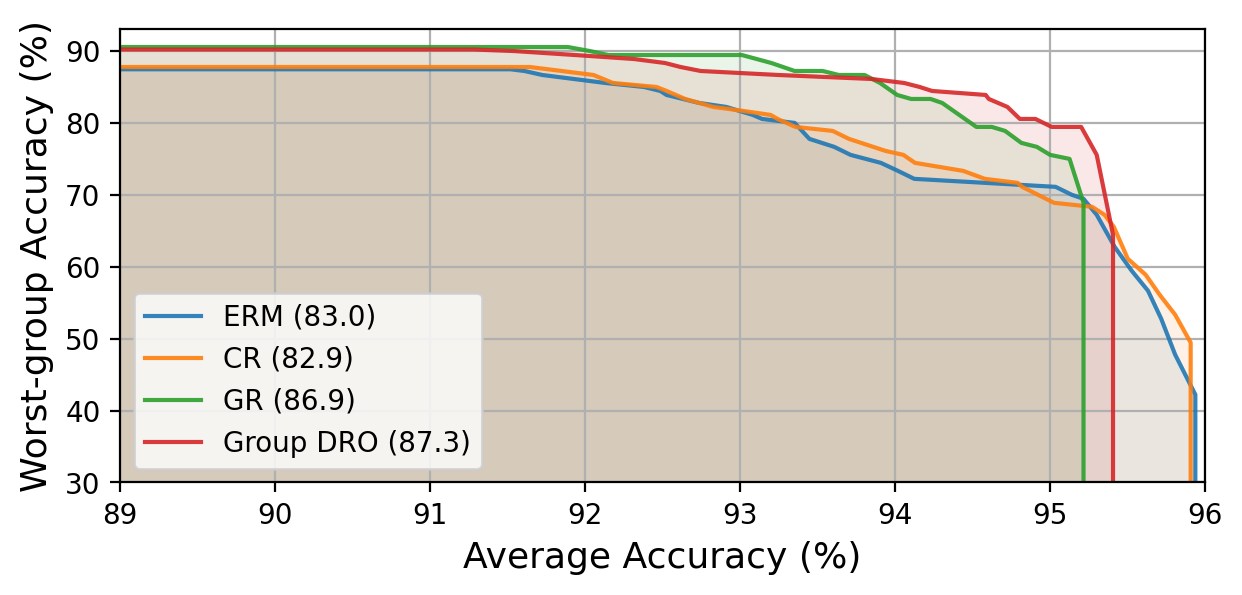}
	\subcaption{Worst-group accuracy}
	\label{fig:worst_curve_all_pareto_celeba}
	\end{subfigure} 
	\hspace{0.5cm}
	    \begin{subfigure}[m]{0.47\linewidth}
    	\includegraphics[width=\linewidth]{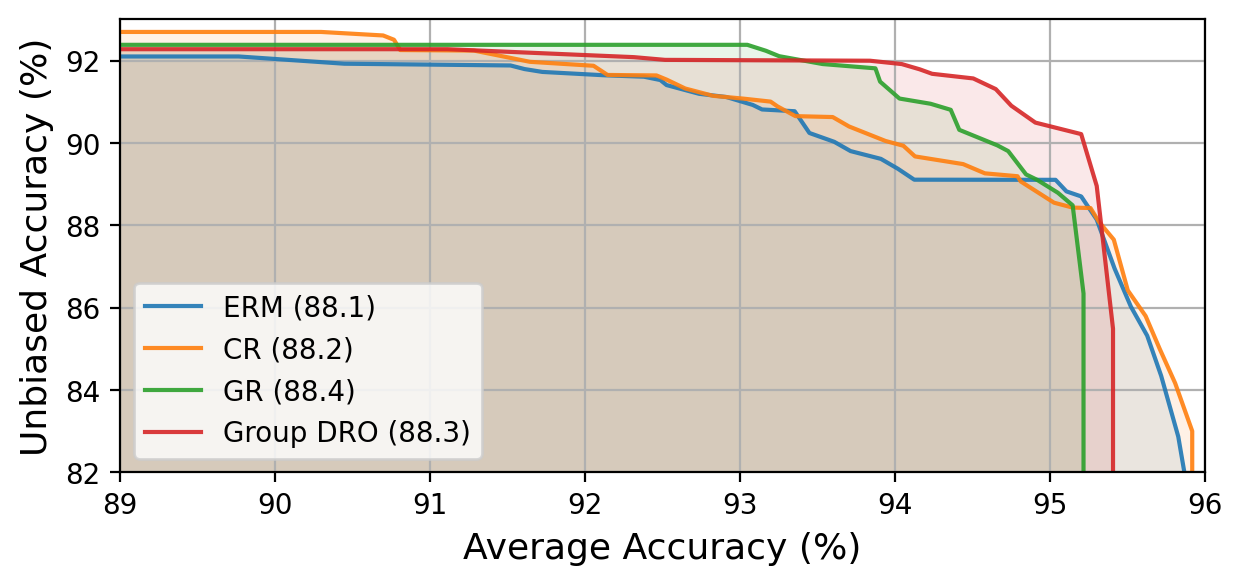}
	\subcaption{Unbiased accuracy}
	\label{fig:unbias_curve_all_pareto_celeba}
	\end{subfigure}
    \caption{The robust-average accuracy trade-off curves ((a), (b)) and their corresponding robust coverage curves ((c), (d)), respectively, on the CelebA dataset.
    The curves in (c) and (d) represent the Pareto frontiers of the curves in (a) and (b), respectively.
	In (c) and (d), the numbers in the legend denote the robust coverage, which measures the area under the curve.
    }
    \vspace{-2mm}
    \label{fig:pareto_celeba}
\end{figure}

\paragraph{Robust coverage curve}
Figure~\ref{fig:worst_curve_all_celeba} and~\ref{fig:unbias_curve_all_celeba} are robust-average accuracy trade-off curves while Figure~\ref{fig:worst_curve_all_pareto_celeba} and~\ref{fig:unbias_curve_all_pareto_celeba} are their corresponding robust coverage curves, which represent the Pareto frontiers of Figure~\ref{fig:worst_curve_all_celeba} and~\ref{fig:unbias_curve_all_celeba}, respectively.
The area under the curve in Figure~\ref{fig:worst_curve_all_pareto_celeba} and~\ref{fig:unbias_curve_all_pareto_celeba} indicates the robust coverage of each algorithm.

\paragraph{Scalability}

  \begin{table}[t]
\begin{center}
\caption{Variations of robust scaling methods and their performances tested on the FairFace dataset.
}
\vspace{3mm}
\label{tab:superclass}
 \scalebox{0.8}{
\setlength\tabcolsep{8pt} 
\begin{tabular}{lc|ccc}
\toprule
Method & Cost & Worst-group & Unbiased &  Average  \\
\hline
ERM & -- &15.8 &47.0 & 54.1\\
+ RS (2 super classes) & $\mathcal{O}(n)$ & 18.6& 51.8 & 52.9\\
+ RS (greedy search) & $\mathcal{O}(n)$ & \bf{19.2} & 52.3 & \bf{53.3} \\
+ RS (full grid search) & $\mathcal{O}(n^9)$ & 19.0& \bf{52.8} & 53.1\\
\bottomrule
\end{tabular}
 }
\end{center}
\end{table}

As mentioned in Section~\ref{sec:robust_scaling}, we search for the scaling factor of each class in a greedy manner.
Hence, the time complexity increases linearly with respect to the number of classes instead of the exponential growth with the full grid search; even with 1,000 classes, the whole process takes less than a few minutes in practice, which is negligible compared to the model training time.
Moreover, we can reduce the computational cost even further by introducing the superclass concept and allocating a single scaling factor for each superclass.
We compare three different options---greedy search, superclass-level search, and full grid search---on the FairFace dataset~\cite{FairFace} with 9 classes. 
Table~\ref{tab:superclass} shows that the greedy search is as competitive as the full grid search despite the time complexity reduction by several orders of magnitude and the superclass-level search is also effective to reduce cost with competitive accuracies.
Note that the superclasses are identified by the feature similarity of class signatures.

 \paragraph {Additional Results}
 
Table~\ref{tab:all_exp} presents full experimental results on the CelebA and Waterbirds datasets, which supplement Table~\ref{tab:celebA} and~\ref{tab:waterbirds}.
We test our robust scaling strategies (RS, IRS) with two scenarios, each aimed at maximizing worst-group or average accuracies, respectively, where each target metric is marked in blue in the tables.

  \begin{table*}[t]
\begin{center}
\caption{Results of our robust scaling methods on top of various baselines on the CelebA dataset, which supplement Table~\ref{tab:celebA}.
\textcolor{blue}{Blue} color denotes the target metric that the robust scaling aims to maximize.
Compared to RS, IRS improves the overall trade-off.
}
\label{tab:all_exp}
 \scalebox{0.78}{
 \hspace{-0.4cm}
\setlength\tabcolsep{6pt} 
\begin{tabular}{l|cc|ccc|ccc}
\toprule
 & \multicolumn{2}{c|}{Robust Coverage} & \multicolumn{3}{c|}{Accuracy (\%)} & \multicolumn{3}{c}{Accuracy (\%)} \\
Method & Worst-group & Unbiased & \textcolor{blue}{Worst-group} & {Unbiased} &  Average & Worst-group & Unbiased &  \textcolor{blue}{Average}  \\
\hline
 ERM  & - & - & 34.5 (6.1) &77.7 (1.8) &\textbf{95.5 (0.4)}   & 34.5 (6.1) &77.7 (1.8) &95.5 (0.4) \\
ERM + RS  &83.0 (0.7) &88.1 (0.5)&82.1 (3.7) &91.1 (0.6) &92.2 (1.3)   &\textbf{45.0 (7.4)} &\textbf{81.7 (1.8)} &\textbf{95.8 (0.2)} \\
ERM + IRS     &\textbf{83.4 (0.1)} & \textbf{88.4 (0.4)} & \textbf{87.2 (2.0)} & \textbf{91.7 (0.2)} &91.5 (0.8)  &44.1 (4.2) &81.3( 0.8) &\textbf{95.8 (0.1)} \\
\cdashline{1-9}
CR & - & - &70.6 (6.0) &88.7 (1.2) &\textbf{94.2 (0.7)}   &\textbf{70.6 (6.0)} &\textbf{88.7 (1.2)} &94.2 (0.7) \\
CR + RS  &82.9 (0.5) &88.2 (0.3) &82.7 (5.2) &91.0 (1.0) &91.7 (1.3)   &48.5 (8.9) &82.5 (2.2) & \textbf{95.8 (0.1)}\\
CR + IRS & \textbf{83.6 (1.1)} &\textbf{88.6 (0.5)} &\textbf{84.8 (1.5)} & \textbf{91.3 (0.4)} & 90.7 (1.3)   &48.8 (9.1) &82.7 (2.4) &\textbf{95.8 (0.1)} \\
\cdashline{1-9}
SUBY & - & - &65.7 (3.9) &87.5 (0.9) &\textbf{94.5 (0.7)}  & \textbf{65.7 (3.9)} & \textbf{87.5 (0.9)} &{94.5 (0.7)}  \\
SUBY + RS  & 81.5 (1.0)  & 87.4 (0.1) & 80.8 (2.9) & 90.5 (0.8) & 91.1 (1.7) & 45.4 (6.7) & 81.4 (2.0) & \textbf{95.5 (0.0)} \\
SUBY + IRS & \textbf{82.3 (1.1)} &\textbf{87.8 (0.2)} &\textbf{82.3 (2.0)} & \textbf{90.8 (0.8)} & 90.7 (1.9)  & 46.0 (6.9) & 81.5 (2.1) & \textbf{95.5 (0.1)}  \\
\cdashline{1-9}
SUBG & - & - &87.8 (1.2) &90.4 (1.2) &\textbf{91.9 (0.3)}   & \textbf{87.8 (1.2)} &\textbf{90.4 (1.2)} &{91.9 (0.3)} \\
SUBG + RS  & 83.6 (1.6) & 87.5 (0.7) & 88.3 (0.7) & 90.9 (0.5) & 90.6 (1.0) & 67.8 (6.5) & 85.2 (2.0) & {93.9 (0.2)} \\
SUBG + IRS & \textbf{84.5 (0.8)} &\textbf{87.9 (0.1)} &\textbf{88.7 (0.6)} & \textbf{91.0 (0.3)} & 90.6 (0.8) & 68.5 (6.5) & 85.5 (1.9) & \textbf{94.0 (0.2)} \\
\cdashline{1-9}
GR & - & - & 88.6 (1.9) &92.0 (0.4) &\textbf{92.9 (0.8)}  & \textbf{88.6 (1.9)} & \textbf{92.0 (0.4)} &{92.9 (0.8)}  \\
GR + RS  &{86.9 (0.4)} &{88.4 (0.2)} & \textbf{90.0 (1.6)} & {92.4 (0.5)} & 92.5 (0.5) &  66.5 (0.3) & 85.4 (0.4) & {93.8 (0.4)} \\
GR + IRS & \textbf{87.0 (0.2)} & \textbf{88.6 (0.2)} & \textbf{90.0 (2.3)} & \textbf{92.6 (0.6)} & 92.5 (0.4) & 62.0 (5.3) & 84.5 (0.7) & \textbf{94.2 (0.3)} \\
\cdashline{1-9}
GroupDRO & - & - &  88.4 (2.3) & 92.0 (0.4) & {93.2 (0.8)}  &  \textbf{88.4 (2.3)} & \textbf{92.0 (0.4)} & {93.2 (0.8)}  \\
GroupDRO + RS  &{87.3 (0.2)} & {88.3 (0.2)} & 89.7 (1.2) & 92.3 (0.1) &\textbf{93.7 (0.5)} & 64.9 (3.3) & 85.1 (0.7)& 93.9 (0.3)\\
GroupDRO + IRS & \textbf{87.5 (0.4)} & \textbf{88.4 (0.2)}& \textbf{90.0 (2.3)} & \textbf{92.6 (0.6)} & 93.5 (0.4) & 60.4 (5.4) & 84.4 (0.6) & \textbf{94.7 (0.3)} \\
\bottomrule
\end{tabular}
}
 \vspace{-0.2cm}
\end{center}

\end{table*}

  \begin{table*}[t]
\begin{center}
\caption{Results of our robust scaling methods on top of various baselines on the Waterbirds dataset, which supplement Table~\ref{tab:waterbirds}.
\textcolor{blue}{Blue} color denotes the target metric that the robust scaling aims to maximize.
Compared to RS, IRS improves the overall trade-off.
}
\label{tab:all_exp}
 \scalebox{0.78}{
 \hspace{-0.4cm}
\setlength\tabcolsep{6pt} 
\begin{tabular}{l|cc|ccc|ccc}
\toprule
  & \multicolumn{2}{c|}{Robust Coverage} & \multicolumn{3}{c|}{Accuracy (\%)} & \multicolumn{3}{c}{Accuracy (\%)} \\
Method & Worst-group & Unbiased & \textcolor{blue}{Worst-group} & {Unbiased} &  Average & Worst-group & Unbiased &  \textcolor{blue}{Average}  \\
\hline
ERM     & - & - &76.3 (0.8) &89.4 (0.6) &\textbf{97.2 (0.2)}   &76.3 (0.8) &89.4 (0.6) &97.2 (0.2) \\
ERM + RS      & 76.1 (0.4) & 82.6 (0.3)  &81.6 (1.9) &89.8 (0.5) &\textbf{97.2 (0.2)}  &\textbf{79.1 (2.7)} & \textbf{89.7 (0.6)}&{97.5 (0.1)} \\
ERM + IRS     &\textbf{83.4 (1.1)} &\textbf{86.9 (0.4)} &\textbf{89.3 (0.5)} &\textbf{92.7 (0.4)} &{94.1 (0.3)}  &{77.6 (7.0)} &{89.6 (1.1)} &\textbf{97.5 (0.3)} \\
\cdashline{1-9}
CR   & - & -  &76.1 (0.7) &89.1 (0.7) &\textbf{97.1 (0.5)}  &76.1 (0.7) &89.1 (0.7) &{97.1 (0.3)} \\
CR + RS   &73.6 (2.3) &82.0 (1.5)  &79.4 (2.4) &89.4 (1.0) &96.8(0.8)  & 76.4(1.5)& \textbf{89.3 (0.8)}&\textbf{97.5 (0.3)} \\
CR + IRS     & \textbf{84.2 (2.5)} & \textbf{88.3 (1.0)} & \textbf{88.2 (2.7)} & \textbf{92. 1(0.7)} & {95.7 (1.1)}  &\textbf{77.3 (4.7)} & 88.6( 1.2)& 97.4 (0.2) \\
\cdashline{1-9}
SUBY   & - & -  & 72.8 (4.1) & 84.9 (0.4) & {93.8 (1.5)} & 72.8 (4.1) & 84.9 (0.4) & {93.8(1.5)} \\
SUBY + RS   &72.5 (1.0) &81.2 (1.4) & 75.9 (4.4) & 86.3 (0.9) & \textbf{95.2 (1.4)} & 70.7 (5.8) & 85.4 (1.6) & {95.5 (0.2)}  \\
SUBY + IRS     & \textbf{78.8 (2.7)} & \textbf{85.9 (1.0)}  & \textbf{82.1 (4.0)} & \textbf{89.1 (0.9)} & 92.6 (2.2) & \textbf{74.1 (4.1)} & \textbf{86.3 (0.9)} & \textbf{96.2 (0.6)} \\
\cdashline{1-9}
SUBG   & - & - & 86.5 (0.9) & 88.2 (1.2) & 87.3 (1.1) & \textbf{86.5 (0.9)} & \textbf{88.2 (1.2)} & 87.3 (1.1) \\
SUBG + RS   & 80.6 (2.0) & 82.3 (2.0) & 87.1 (0.7) & \textbf{88.5 (1.2)} & \textbf{87.9 (1.1)} & 74.0 (5.6) & 85.9 (2.8) & 91.3 (0.4) \\
SUBG + IRS   & \textbf{82.2 (0.8)} & \textbf{84.1 (0.8)} & \textbf{87.3 (1.3)} & 88.2 (1.2) & 87.6 (1.2) & 70.2 (1.6) & 84.5 (1.0) & \textbf{93.5 (0.4)} \\
\cdashline{1-9}
GR   & - & -  &86.1 (1.3) &89.3 (0.9) &\textbf{95.1 (1.3)}  &\textbf{86.1 (1.3)} &89.3 (0.9) &{95.1 (1.3)}  \\
GR + RS  &{83.7 (0.3)} & {86.8 (0.7)}&\textbf{89.3 (1.3)} &{92.0 (0.7)} & 93.1 (3.2) & 82.2 (1.3) & \textbf{90.8 (0.5)} &{95.4 (1.3)} \\
GR + IRS  & \textbf{84.8 (1.7)} &\textbf{87.4 (0.4)} &89.1 (0.8) & \textbf{92.2 (1.0)} & 92.9 (2.1) &  82.1 (1.4) & 90.5 (0.7) & \textbf{95.6 (0.8)} \\
\cdashline{1-9}
GroupDRO   & - & -  &88.0 (1.0) &92.5 (0.9) &{95.8 (1.8)} &\textbf{88.0 (1.0)} &\textbf{92.5 (0.9)} &{95.8 (1.8)} \\
GroupDRO + RS   & {83.4 (1.1)}&{87.4 (1.4)} &{89.1 (1.7)} &{92.7 (0.8)} & \textbf{96.4 (1.5)} & 80.9 (4.4) & 91.3 (1.0) & \textbf{97.1 (0.3)} \\
GroupDRO + IRS     & \textbf{86.3 (2.3)} & \textbf{90.1 (2.6)} & \textbf{90.8 (1.3)} & \textbf{93.9 (0.2)} & {96.0 (0.6)} & 83.2 (1.7) & 91.5 (0.8) & \textbf{97.1 (0.4)}  \\
\bottomrule
\end{tabular}
}
 \vspace{-0.2cm}
\end{center}

\end{table*}

\section{Discussion}

\paragraph{Limitation}
Although our framework is simple yet effective for improving target metrics with no extra training, it does not learn debiased representations as it is a post-processing method. 
However, this suggests that existing training approaches may also not actually learn debiased representations, but rather focus on prediction adjustment for group robustness in terms of robust accuracy.
From this point of view, our comprehensive measurement enables a more accurate and fairer evaluation of base algorithms, considering the full landscape of trade-off curve.

\end{document}